\documentclass{article} 
\usepackage[preprint]{goodfire/goodfire}

\usepackage{url}
\usepackage{booktabs}
\usepackage{amsmath}
\usepackage{amssymb}
\usepackage{mathtools}
\usepackage{amsthm}
\usepackage{enumitem}
\usepackage{microtype}
\usepackage{graphicx}
\usepackage{subcaption}
\usepackage{multirow}
\usepackage{wrapfig}
\usepackage[bottom]{footmisc}

\usepackage{hyperref}
\usepackage{epigraph}
\usepackage{etoolbox}
\usepackage[most]{tcolorbox}
\usepackage{macro}

\usepackage[capitalize,noabbrev]{cleveref}

\usepackage[color=red!20,textsize=tiny]{todonotes}

\usepackage{lineno}

\hypersetup{colorlinks=true, citecolor=primary, linkcolor=primary, urlcolor=primary}

\providecommand{\aut}[1]{\textbf{#1}}
\providecommand{\af}[1]{{\small #1}}

\providecommand{\afn}[1]{\textcolor{primary}{$^{#1}$}}

\newcommand{\costar}{\textcolor{secondary}{\boldsymbol{\star}}}
\newcommand{\colead}{\textcolor{secondary}{\boldsymbol{\dagger}}}

\newcommand{\goodfiremark}{\raisebox{0.5pt}{\hspace{0.5mm}\includegraphics[height=6pt]{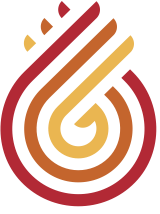}}}

\newcommand{\goodfireaff}{%
  \includegraphics[height=14pt]{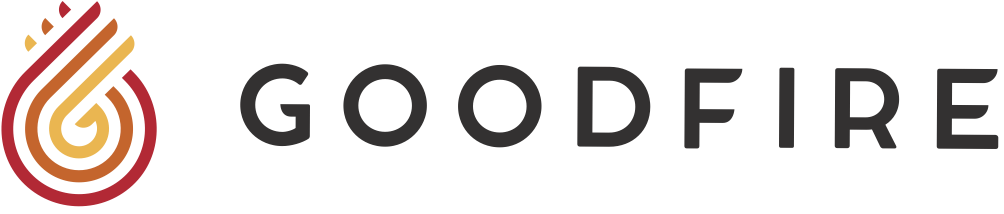}
}

\newcommand{\authorentry}[2]{\aut{#1}\afn{#2}}
\newcommand{\afflabel}[2]{\afn{#1}\af{#2}}
\newcommand{\authorsep}{\quad}
\newcommand{\repolink}[1]{%
    {\small
      \href{#1}{\raisebox{-2.8pt}{\includegraphics[height=10pt]{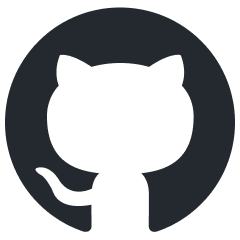}}%
      ~\textcolor{secondary}{\nolinkurl{#1}}}
    }%
}

\newtoggle{goodfireonly}
\togglefalse{goodfireonly}  
\newcommand{\extline}[1]{\iftoggle{goodfireonly}{}{#1}}
\newcommand{\extaff}[1]{\iftoggle{goodfireonly}{}{#1}}

\newcommand{\paperauthors}{%
\authorentry{Daniel Wurgaft}{\costar\goodfiremark\extaff{,a}} \authorsep
\authorentry{Can Rager}{\costar\goodfiremark\extaff{,b}} \authorsep
\authorentry{Matthew Kowal}{\costar\goodfiremark} \authorsep 
\authorentry{Vasudev Shyam}{\goodfiremark} \vspace{2pt} \\
\authorentry{Sheridan Feucht}{\goodfiremark\extaff{,c}} \authorsep
\authorentry{Usha Bhalla}{\goodfiremark\extaff{,d}} \authorsep
\authorentry{Tal Haklay}{\goodfiremark} \authorsep
\authorentry{Eric Bigelow}{\goodfiremark\extaff{,e}} \authorsep
\authorentry{Raphael Sarfati}{\goodfiremark} \authorsep \vspace{2pt} \\ 
\authorentry{Thomas McGrath}{\goodfiremark} \authorsep
\authorentry{Owen Lewis}{\goodfiremark} \authorsep
\authorentry{Jack Merullo} {\goodfiremark} \authorsep
\authorentry{Noah D. Goodman}{\colead, a} \vspace{2pt} \\ 
\authorentry{Thomas Fel}{\colead\goodfiremark} \authorsep
\authorentry{Atticus Geiger}{\colead\goodfiremark} \authorsep
\authorentry{Ekdeep Singh Lubana}{\colead\goodfiremark}
\vspace{4pt}\\
\textcolor{secondary}{$^{\boldsymbol{\star}}$}\af{Equal contribution} \authorsep
\vspace{3mm}
\textcolor{secondary}{$^{\boldsymbol{\dagger}}$}\af{Equal senior contribution} \\
\goodfireaff \\
\extline{%
  \afflabel{a}{Stanford University} \authorsep
  \afflabel{b}{University College London} \authorsep
  \afflabel{c}{Northeastern University} \authorsep\\
  \afflabel{d}{Harvard University} \authorsep
  \afflabel{e}{Technion IIT}
}
\vspace{3mm}\\
\repolink{https://github.com/goodfire-ai/causalab/tree/manifold_steering}
\vspace{-4mm}
}
\author{\paperauthors}

\title{Manifold Steering Reveals the Shared Geometry of Neural Network Representation and Behavior }

\begin{document}

\maketitle

\vspace{-4mm}
\begin{abstract}
Neural representations carry rich geometric structure; but does that structure causally shape behavior? 
To address this question, we intervene along paths through activation space defined by different geometries, and measure the behavioral trajectories they induce.
In particular, we test whether interventions that respect the geometry of activation space will yield behaviors close to those the model exhibits naturally.
Concretely, we first fit an activation manifold $\mathcal{M}_h$ to representations and a behavior manifold $\mathcal{M}_y$ to  output probability distributions.
We then test the link $\mathcal{M}_h \leftrightarrow \mathcal{M}_y$ via interventions: we find that steering along $\mathcal{M}_h$, which we term \textit{manifold steering}, yields behavioral trajectories that follow $\mathcal{M}_y$, while linear steering---which assumes a Euclidean geometry---cuts through off-manifold regions and hence produces unnatural outputs. 
Moreover, optimizing interventions in activation space to produce paths along $\mathcal{M}_y$ recovers activation trajectories that trace the curvature of $\mathcal{M}_h$.
We demonstrate this bidirectional relationship between the geometry of representation and behavior across tasks and modalities. 
In language models, we use reasoning tasks with cyclic and sequential geometries as well as in-context learning tasks with more complex graph geometries. In a video world model, we use a task with geometry corresponding to physical dynamics.
Overall, our work shows that geometry in neural representation is not merely incidental, but is in fact the proper object for enabling principled control via intervention on internals. 
This recasts the core problem of steering from finding the right \textit{direction} to finding the right \textit{geometry}.

\end{abstract}

\section{Introduction}
\label{sec:intro}
A plethora of geometric structures have been documented in neural network representations \citep{modell2025originsrepresentationmanifoldslarge, park2025iclr, kozlowski2025semanticstructurelargelanguage, shai2024transformersrepresentbeliefstate, pearce2025tree, gurnee2026models}. 
Recent literature has begun to identify the origins of these structures by attributing them back to data statistics shaped by conceptual structure~\citep{karkada2026symmetrylanguagestatisticsshapes, prieto2026correlations, park2025iclr, merullo2025linearrepresentationspretrainingdata}.
However, we have barely begun to understand what causal role these geometric structures play in a model's computation (cf. \citealt{engels2024not, kantamnenihelix, csordas2024recurrent,  sarfati2026shapebeliefsgeometrydynamics}). 
We address this question by intervening on model activations under different geometric assumptions and measuring the effect on behavior.

Currently, it is common for activation-based intervention methods to assume a Euclidean geometry for activation space, where steering is performed by adding a \textit{steering vector} to model activations with a scalar that modulates intervention strength \citep{bau2018identifying, subramani-etal-2022-extracting, marks2024geometrytruthemergentlinear, panickssery2024steeringllama2contrastive, turner2024steeringlanguagemodelsactivation, li2024inferenceintervention, rimsky2024contastive, chen2025persona}. 
This approach is motivated by the \textit{linear representation hypothesis} (LRH), which posits that neural activations can be decomposed into atomic concepts encoded along single (approximately) orthogonal directions \citep{Smolenksy, park2023linear, elhage2022toy}.  However, linear steering often produces degraded fluency, diversity collapse, and unstable off-target behavior~\citep{wu2025axbenchsteeringllmssimple, da2025steering,bigelow2025beliefdynamicsrevealdual,tan2024analysing, hao2025patterns, bhalla2024towards, pres2024towards}, which suggests the assumed Euclidean geometry is inappropriate.

\begin{figure}[t]
    \centering
    \vspace{-10pt}
    \includegraphics[width=\textwidth]{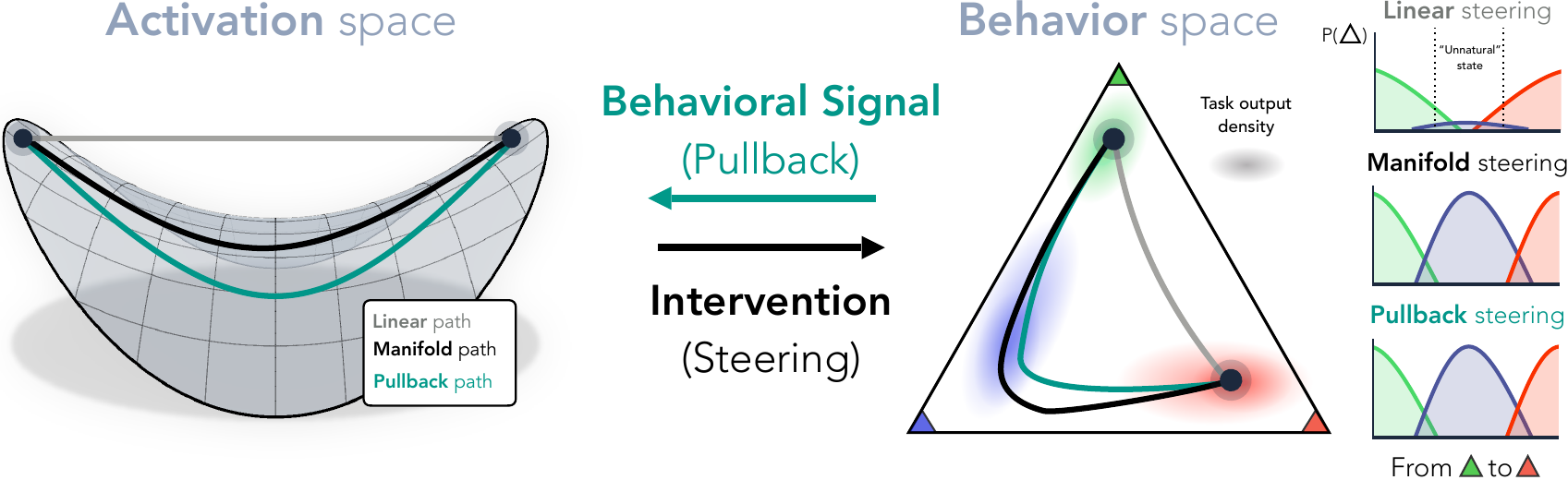}
\caption{\textbf{How do different geometries of activation space modulate behavior?}
We illustrate paths through activation space (left), each defined by a different geometry. Interventions along paths in activation space induce paths in behavior space (right, illustrated on a three-concept probability simplex).
\textbf{Euclidean:} the standard approach of linear steering assumes a flat geometry and interventions follow a straight line. Such paths may cut across the activation manifold, yielding \textit{unnatural} behavioral trajectories that pass through off-manifold regions of behavior space.
\textbf{Density geometry:} a density-based metric whose geodesics follow the intrinsic geometry of a fitted activation manifold, yielding more natural transitions in behavior space.
\textbf{Pullback geometry:} a behavior-aware metric obtained by ``pulling back'' behavior-space geometry into activation space, yielding paths that follow the manifold of natural (unintervened) output distributions.
Overall, we argue that geometric structure in neural representations encodes the conceptual space a model is reasoning over, which in turn constrains its output behavior. Hence, manifolds in activation and behavior space are two images of the same underlying structure, and so we expect the density and pullback geometries to coincide.
}
\label{fig:intro-fig}

\end{figure}

In this work, we advance the hypothesis that \textit{representation geometry} provides a blueprint for effective steering that will overcome the limitations of the linear approach. 
Steering is fundamentally about how internal representations control behavior, so to test this hypothesis we must study not only paths through activation space, but also the behavioral trajectories induced by interventions along these paths. 
Successful steering will produce trajectories that are in line with the model's  \textit{natural} (unintervened) output distribution.
If we are right, then interventions that respect the geometry of internal representations and interventions that respect the geometry of behavior will be one and the same. 
Motivated by this, we make the following contributions in this work.

\begin{itemize}[leftmargin=*]
    \item \textbf{Uncovering isometric geometries in neural network representation and behavior.}
    We use tasks where models output a distribution over a set of concepts with known structure. In each task, we fit an \emph{activation manifold} $\mathcal{M}_h$ to internal representations and a \emph{behavior manifold} $\mathcal{M}_y$ to model outputs (probability distributions over task-relevant concepts). We show the two geometries are tightly interlinked via a scaled isometry relation: geodesic distances on $\mathcal{M}_h$ align closely with those on $\mathcal{M}_y$, and neither match Euclidean distances.

    \item \textbf{Validating the causal role of representation geometry.}
    We perform geometry-aware steering experiments and compare against the baseline of linear steering (see Fig.~\ref{fig:intro-fig}). 
    We show linear steering cuts through low-density regions of behavior space and passes through unnatural intermediate distributions; meanwhile, steering along the activation manifold $\mathcal{M}_h$ yields behavioral trajectories that follow $\mathcal{M}_y$ closely. 
    In fact, optimizing for paths along $\mathcal{M}_y$ recovers activation trajectories that trace the curvature of $\mathcal{M}_h$, further tightening the link between activation geometry and behavior.
    
    \item \textbf{A theoretical framework for geometry-aware steering.} Building on the results above, we formulate steering as a problem of choosing the right \textit{geometry} for activation space, rather than the right \textit{direction}. 
    In particular, we argue steering can be defined as the problem of finding a geodesic connecting two points under different activation-space metrics: linear steering assumes a flat metric (Euclidean geometry), steering along the activation manifold uses a metric derived from natural activations, and steering optimized to follow the behavior manifold uses a metric derived from natural behaviors.
\end{itemize}

We demonstrate these findings hold across modalities and tasks.
In large language models, we test geometries from cyclic concepts (weekdays, months; \citealt{engels2024not, modell2025originsrepresentationmanifoldslarge}), sequential concepts (ages, letters), and multi-dimensional graph structures learned in context \citep{park2025iclr}.
In a video world model, we test a geometry of physical position in a simulated environment (mountain car; \citealt{moore1990efficient, towers2024gymnasium}).
Together, these findings provide evidence for the posited account, and support steering along neural manifolds as the principled form of activation-based intervention.

\section{The Geometry of Representation and Behavior}
\label{sec:geometry}
\newcommand{\behaviorsp}{\mathcal{Y}}
\subsection{Setup}
\paragraph{Running example.} 
We will explicate our framework and empirical methods using a running example where a language model is required to reason about the days of the week \citep{engels2024not}. Specifically, we consider prompts of the form: \texttt{What day is $k$ days after $z$?} with $z \in \mathcal{Z} = \{\mathrm{Mon}, \mathrm{Tue}, \ldots, \mathrm{Sun}\}$ and $k \in \{1, \dots, 7\}$. Given such a prompt, the LM outputs a probability distribution over all possible tokens. 

\paragraph{Concept geometry.}
We draw inspiration from work on \textit{conceptual spaces} in cognitive science, where conceptual domains, e.g., days of the week, are geometrically enriched with a metric such that distances between points encode similarity and guide patterns of inference (\citealt{shepard1987toward, Gardenfors:2000, tenenbaum2001generalization, bellmund2018navigating}; see \citealt{fel2025into, lubana2025priors, yocum2025neural, modell2025originsrepresentationmanifoldslarge} for related work in interpretability).  For example, the days of the week $\mathcal{Z}$ may be organized in a cyclic structure that is captured by a \textit{metric} $d_{\mathcal{Z}}$ measuring temporal distance between days, e.g., neighboring days are closer together. Indeed, when humans mistakenly report the current day, they most often confuse it with its neighboring days \citep{ellis2015mental}.

\cite{karkada2026symmetrylanguagestatisticsshapes} and \cite{prieto2026correlations} show that similarity structure between days of the week is reflected in the statistics of training data, which in turn shape the geometry of internal representations \citep{engels2024not, park2025, model2025, prieto2026correlations}. 
We hypothesize that a model's output
distributions over $\mathcal{Z}$ are similarly shaped by $d_{\mathcal{Z}}$: e.g., when asked \texttt{ What day is four days after Monday?}, the model concentrates mass on Friday and spreads the remainder onto nearby days like Thursday and Saturday. 

\paragraph{Notation.}
We work with two spaces: the activation space $\mathcal{A} = \mathbb{R}^n$, and the behavior space $\behaviorsp = \Delta^{|\mathcal{Z}|}$, which is the open probability simplex\footnote{We require the open simplex
$\{\bm{p} \in \mathbb{R}^{|\mathcal{Z}|}_{>0} : \sum_i p_i = 1\}$, i.e., strictly positive entries. The closed simplex, which includes faces where some $p_i = 0$, has boundary and corners and is not a smooth manifold.} over the conceptual domain $\mathcal{Z}$, with an additional `other' class for off-concept probability mass.
For an input $x$, let $\bm{p}(x) \in \behaviorsp$ denote the model's output distribution over $\mathcal{Z}$, given by restricting the full vocabulary distribution of the model to the tokens in $\mathcal{Z}$, in addition to the `other' class for remaining probability mass. Let $\bm h(x) \in \mathcal{A}$ denote an activation vector of interest for input $x$.

For a class of input queries that share the same answer, e.g., \texttt{What is two days after Monday?} and \texttt{What is three days after Sunday?}, we average the hidden activations and output distributions to produce ``activation centroids'' and ``behavior centroids'', respectively.

\paragraph{Experimental tasks.} We perform language model experiments on four tasks, two with cyclic conceptual structure and two with sequential conceptual structure. 
The cyclic tasks require reasoning about days of the week and months of the year, e.g., \texttt{What is four months after January?}. 
The sequential tasks require reasoning about letters and ages, e.g., \texttt{What is four letters after m?} or \texttt{Alice is 7, Bob is 5 years older. How old is Bob?}.

\subsection{Fitting the Manifolds}

We fit a smooth manifold within each space to the model's unintervened activations or outputs for a task: $\mathcal{M}_h \subseteq \mathcal{A}$, the \textit{activation manifold}, and $\mathcal{M}_y \subseteq \mathcal{Y}$, the \textit{behavior manifold}. To fit the activation manifold $\mathcal{M}_h$, we reduce activation vectors $\bm h(x)$ to 64 dimensions via PCA, compute ``concept centroids'' (e.g., averaging all 
activations where the correct answer is Wednesday), and fit cubic splines \citep{reinsch1967smoothing} through 
the centroids (see 
App.~\ref{app:fit-mh} for further spline fitting details). To fit the behavior manifold $\mathcal{M}_y$, we follow a similar procedure but first map each centroid from the probability simplex onto Hellinger space via                                              
$p \mapsto \sqrt{p}$. This linearizes                                           
the geometry of the simplex: the Hellinger distance between     
distributions becomes an ordinary Euclidean distance,                                                       
$d_H(p, q) = \tfrac{1}{\sqrt{2}}\|\sqrt{p} - \sqrt{q}\|$, so we can
fit splines and compare distributions with standard Euclidean tools                                         
while still respecting the underlying probabilistic geometry    
\citep{amari2000methods}. Decoded points are squared back to recover                                        
valid distributions (further details, including how we keep the fit
on the sphere, are in App.~\ref{app:fit-my}).  Unless stated otherwise, we use 
Llama 3.1 8B~\citep{touvron2023llama} with activations from layer 28, and visualize manifolds via $3$D PCA. 

\begin{figure}[!t]  
\includegraphics[width=0.95\linewidth]{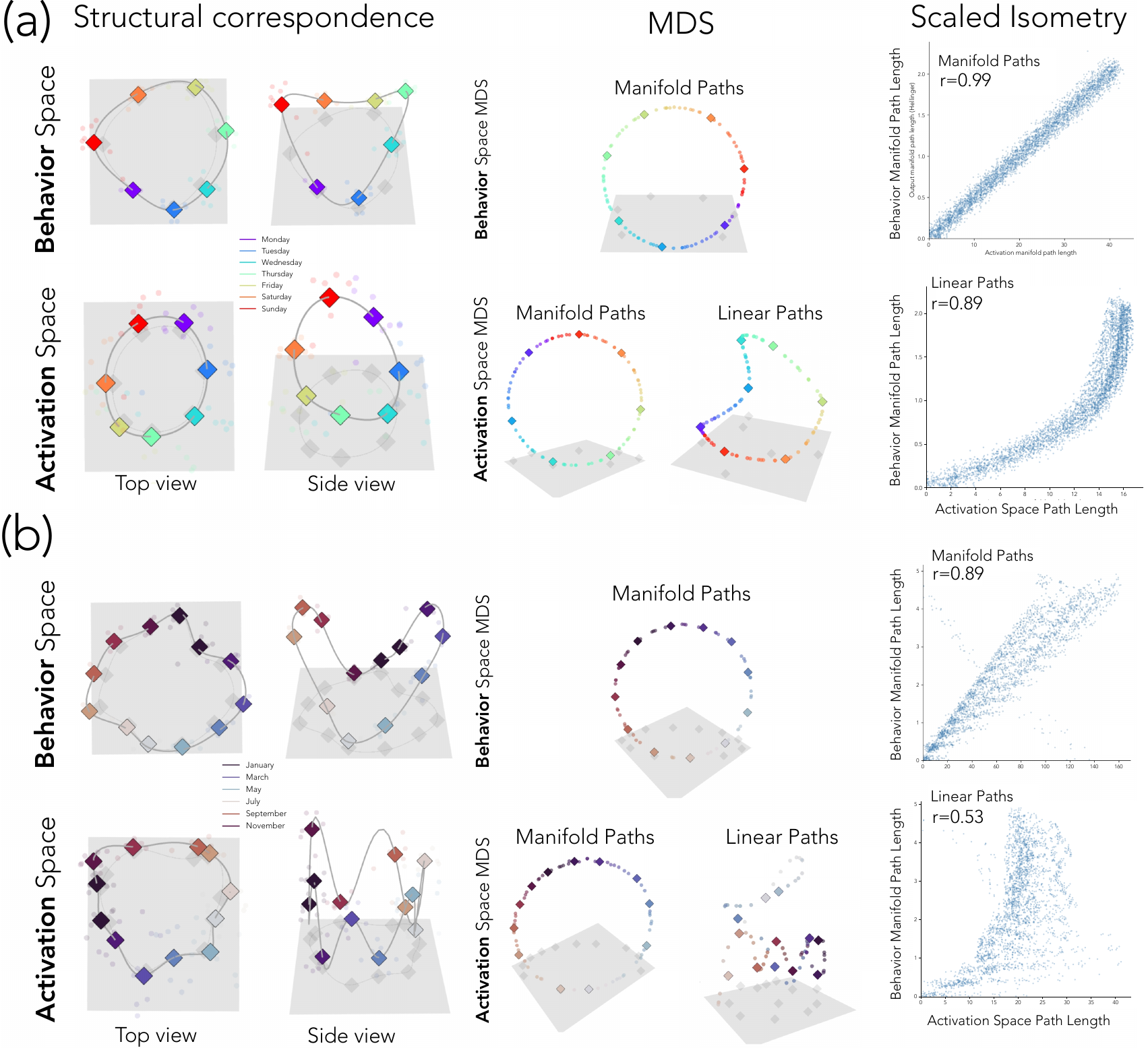}
    \caption{\textbf{Approximate isometry between activation and behavior manifolds for cyclic concepts}. Manifolds (cubic splines) fit to activation and behavior (i.e., output distributions over concept tokens) spaces of Llama 3.1 8B. The weekdays \textbf{(a)} and months \textbf{(b)} tasks consist of simple addition questions such as: \texttt{What is four days after Monday?}. Both activation and behavior manifolds show cyclic structure (PCA visualization shown in left column). Furthermore, on-manifold distances in activation space show strong correlation with on-manifold distances in behavior space (right column), as well as a clear structural match via a multidimensional scaling (MDS) embedding (middle column). In contrast, linear distances in activation space show weaker correlations and warped structures. These results demonstrate an approximate isometry between the activation and behavior space manifolds.}
    \label{fig:shapes-wrapfigabove_below}
    \vspace{-5pt}
\end{figure}

\begin{figure}[!t]  
\includegraphics[width=0.95\linewidth]{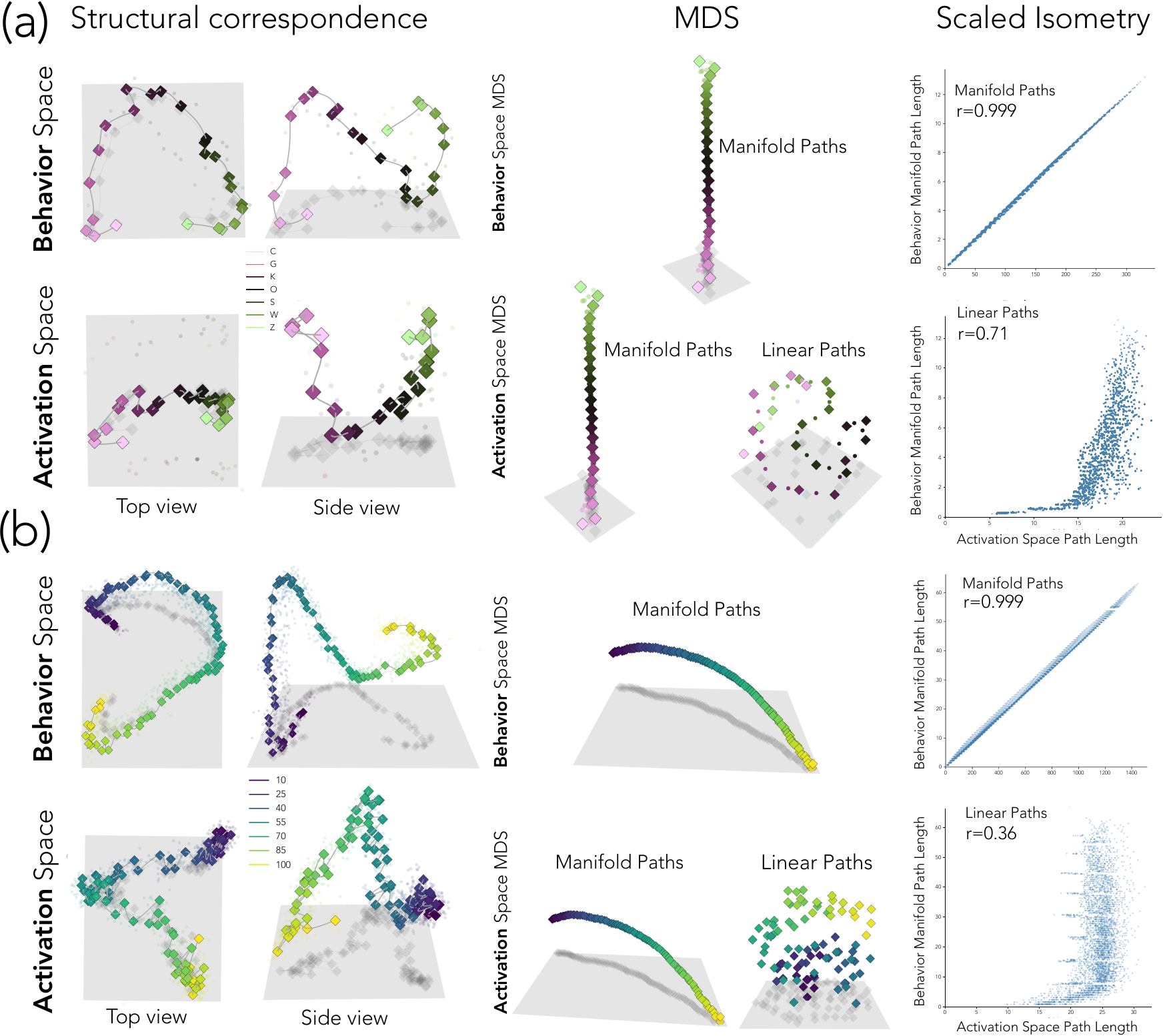}
    \caption{\textbf{Approximate isometry between activation and behavior manifolds for sequential concepts}. Manifolds (cubic splines) fit to activation and behavior (i.e., output distributions over concept tokens) spaces of Llama 3.1 8B. The letters \textbf{(a)} and ages \textbf{(b)} tasks consist of simple addition questions such as: \texttt{What letter comes four letters after M?}. Both activation and behavior manifolds show sequential structure (PCA visualization shown in left column). Furthermore, on-manifold distances in activation space show strong correlation with on-manifold distances in behavior space (right column), as well as a clear match via a multidimensional scaling (MDS) embedding (middle column). In contrast, linear distances in activation space show weaker correlations and warped or incoherent structure. These results demonstrate an approximate isometry between the activation and behavior space manifolds.}
    \label{fig:shapes-wrapfigabove_below_repeat}
    \vspace{-5pt}
\end{figure}

\subsection{Conceptual Structure Appears in Behavior and Activation Space}
We will now begin our investigation into the connection between the activation and behavior manifolds. Before we perform any interventions on internal representations, we 
examine the structural correspondence between these spaces, and measure whether distances along the two manifolds are proportional (a scaled isometry).

We find that both the activation and behavior manifolds recapitulate conceptual structure (see Figs.~\ref{fig:shapes-wrapfigabove_below},~\ref{fig:shapes-wrapfigabove_below_repeat} for visualizations). For example, in the case of days of the week,
the activations and output distributions are arranged in order around a loop, with Monday adjacent to Tuesday and Sunday, and Thursday on the opposite side (Fig.~\ref{fig:shapes-wrapfigabove_below}). The circle representing days of the week in activation space is already known to exist~\citep{engels2024not, 
modell2025originsrepresentationmanifoldslarge, karkada2026symmetrylanguagestatisticsshapes}. 
However, the circle in behavior space is a novel discovery, and results from sharply-peaked output distributions placing most mass on the target concept, with the remainder concentrated on its neighbors. The correspondence is striking; output distributions and internal activations recover the same cyclic ordering.
In contrast, the conceptual structure for the ages and letters task is sequential rather than cyclic, and so both the activations and output distributions for these tasks lie on an open curve (Fig.~\ref{fig:shapes-wrapfigabove_below_repeat}). 

Going beyond qualitative structural correspondence, we wish to examine the mapping between distances along each manifold. We test this by computing pairwise distances between points in both spaces: geodesic distances
$d_{\mathcal{M}_h}(m_i, m_j)$ on the activation manifold, and
geodesic distances $d_{\mathcal{M}_y}(p_i, p_j)$ on
the behavior manifold. We compute geodesic distance on $\mathcal{M}_h$ using cumulative Euclidean distance between points along a geodesic path, and follow the same procedure for geodesic distance on $\mathcal{M}_y$, but using cumulative Hellinger distance (see App.~\ref{app:geodesics} for further details).
The two distances are highly correlated ($r = 0.99$ weekdays, $r = 0.89$
months, $r=.999$ letters, $r=.999$ ages) indicating that $\mathcal{M}_h$ and $\mathcal{M}_y$
are approximately isometric. Meanwhile, linear paths between the same activation-space points correlate less well with $\mathcal{M}_y$ geodesics, with the relationship showing clear non-linear patterns 
($r = 0.89$ weekdays, $r = 0.53$ months, $r=0.71$ letters, $r=0.36$ ages).

This correspondence leads to an intuitive hypothesis. $\mathcal{M}_y$ was fit to unintervened task behavior, so it traces a path through \textit{natural} output distributions for the model. If the $\mathcal{M}_h \leftrightarrow \mathcal{M}_y$ mapping holds, paths along one manifold should track paths along the other. Interventions in activation space that follow $\mathcal{M}_h$ should produce natural trajectories along $\mathcal{M}_y$. Conversely, activation-space paths that are optimized to produce trajectories on $\mathcal{M}_y$ should recover $\mathcal{M}_h$. Next, we test both directions via intervention: representation to behavior ($\mathcal{M}_h \rightarrow \mathcal{M}_y$) and behavior to representation ($\mathcal{M}_h \leftarrow \mathcal{M}_y$).

\section{Connecting Representation and Behavior via Intervention}
\label{sec:phenomenology}
Now that we have established a correlational correspondence between the activation and behavior manifolds across four tasks, we turn to steering interventions for causal evidence. First, we steer along $\mathcal{M}_h$ and measure whether output trajectories follow $\mathcal{M}_y$ (\S\ref{sec:act1}). Second, we optimize interventions on internal representations to produce output distributions that follow $\mathcal{M}_y$ and measure whether the optimized activation trajectory follows $\mathcal{M}_h$ (\S\ref{sec:act2}).

\subsection{Steering Intervention Notation}
The basic intervention operation entails replacing the model's activation at a chosen layer with a target activation, and continuing the forward pass. 
Given a base input $x$ and a target $\bm{h}^\star \in \mathcal{A}$, we write $\bm{p}_{\bm{h} \leftarrow \bm{h}^\star}(x)$ for the resulting output distribution. 
A \emph{steering path} is a curve $\bm{\pi} : [0,1] \to \mathcal{A}$ between endpoints $\bm{h}^\star_0$ and $\bm{h}^\star_1$, inducing a trajectory $\bm{p}_{\bm{h} \leftarrow \bm{\pi}(t)}(x)$ through behavior space $\behaviorsp$. 
The behavioral trajectory will be non-stationary only if the target $\mathbf{h}$ mediates the causal effect from input to output \citep{pearl2001directindirecteffects, vig2020, mueller2024}.

We consider two strategies, both constructed by interpolation between the endpoints; the strategies differ only in the coordinate system in which the interpolation is taken (Fig.~\ref{fig:intro-fig}):
\begin{align}
    \bm{\pi}_{\mathrm{lin}}(t) &= (1{-}t)\, \bm{h}^\star_0 + t\, \bm{h}^\star_1
        & \text{(linear steering);} \label{eq:linear-steering} \\
    \bm{\pi}_{\mathrm{m}}(t) &= \bm{s}\bigl((1{-}t)\, \bm{u}_0 + t\, \bm{u}_1\bigr),
        \quad \bm{u}_i = \bm{s}^{-1}(\bm{h}^\star_i)
        & \text{(manifold steering).} \label{eq:geometric-steering}
\end{align}
In the above, $\bm{s} : \mathbb{R}^k \to \mathcal{A}$ is a \emph{parameterization} of $\mathcal{M}_h$---the map sending $k$-dimensional intrinsic coordinates to the corresponding point on the manifold in the activation space $\mathcal{A}$. 
Linear steering (also known as `diff-in-means steering')~\citep{bau2018identifyingcontrollingimportantneurons, subramani-etal-2022-extracting, turner2023activation} interpolates in $\mathcal{A}$ directly---the standard additive-vector baseline. 
Manifold steering interpolates in the intrinsic coordinates of $\mathcal{M}_h$ and maps the result back through $\bm{s}$, so $\bm{\pi}_{\mathrm{m}}$ stays on the activation manifold $\mathcal{M}_h$ throughout. 
Each strategy thus corresponds to a different choice of geometry on activation space, which we concretize in \S\ref{sec:steering-as-metric}.

\begin{figure}[!t]
    \centering
    \includegraphics[width=1\linewidth]{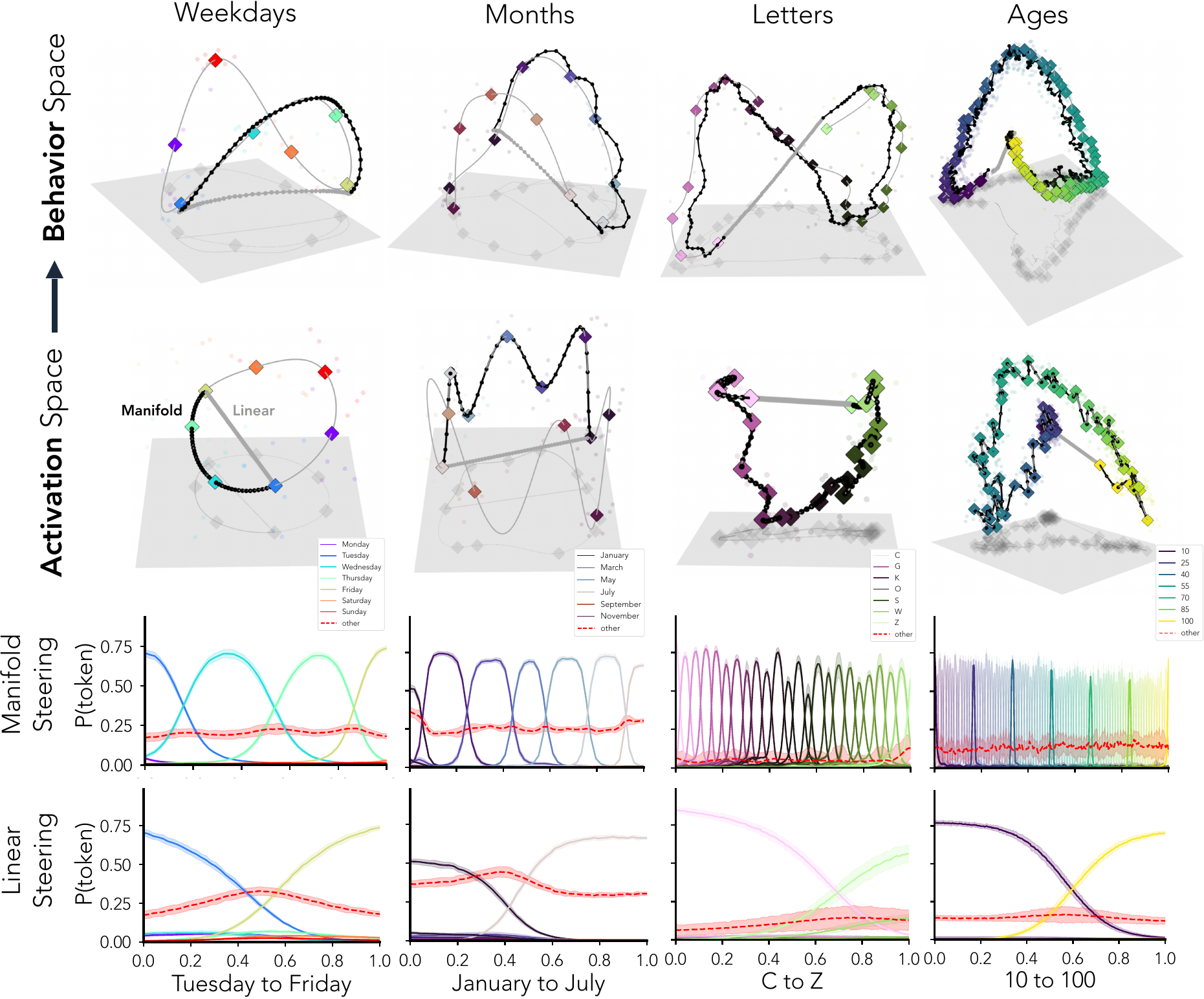}
    \caption{\textbf{Manifold steering yields smooth and ordered behavioral transitions.}
    Using simple addition tasks which require reasoning over structured concepts (e.g., \texttt{What is four days after Monday?}), we compare two steering strategies in activation space: standard linear steering, which takes direct paths, and \textit{manifold steering}, which takes paths along a fitted activation manifold. The bottom panel shows example output paths given by each method. Across four settings, manifold steering produces smooth and ordered output transitions between adjacent concepts. In contrast, linear steering leads to `teleportation' of probability between non-adjacent concepts, and at times results in probability on non-related tokens surpassing any individual concept near the path midpoint. The top panel shows output trajectories in behavior space resulting from manifold steering. We find that steering along the activation-space manifold yields paths that follow the behavior manifold, while linear steering traces paths far from the manifold. Thus, the outputs produced under manifold steering more closely resemble \textit{natural} outputs produced without intervention.}
    \label{fig:manifold-steering-main}
\end{figure}

\subsection{Steering Along the Activation Manifold Follows the Behavior Manifold}
\label{sec:act1}
For every pair of start and end values, e.g., from Tuesday to Friday, we steer from the start centroid to the end centroid in activation space using manifold and linear steering with $K = 50$ intervention points along each path. We report the average trajectory in behavior space over a set of 16 prompts sampled randomly from the task's input distribution (Fig.~\ref{fig:manifold-steering-main}, see App. ~\ref{app:steering} for further experimental details). We find that manifold steering produces \emph{smooth} and \emph{ordered} behavioral transitions: probability mass shifts steadily through adjacent values of a concept---from Monday to Tuesday to Wednesday to Thursday---while linear steering instead exhibits \emph{`teleportation'}: mass jumps between non-adjacent concepts as the straight line cuts through the manifold's 
interior.

This qualitative evidence is encouraging, but we have yet to examine our key hypothesis that interventions along the activation manifold $\mathcal{M}_h$ produce \textit{natural} output trajectories that follow $\mathcal{M}_y$. This would mean that outputs produced under manifold steering resemble those produced without intervention.
We quantify this via an ``energy function'', as described next, under which a natural trajectory is one of low cumulative energy as defined by $\mathcal{M}_y$.

\label{sec:naturalness}
\paragraph{An Energy-based View of Naturalness.} Energy functions have a long history in machine learning as a way to measure plausibility under a model~\citep{hopfield1982neural, lecun2006tutorial}. 
These functions assign low values for likely states and
high values for unlikely ones, with the standard correspondence
$E(\bm{x}) \propto -\log p(\bm{x})$ giving energy the interpretation of an unnormalized log-density~\citep{hopfield1982neural,lecun2006tutorial,grathwohl2019your,song2021train,bethune2025follow}. We adopt the same view here. 
The model's output distributions on unintervened forward passes trace out a low-energy region of behavior space (approximately captured by the manifold $\mathcal{M}_y$) and a steering trajectory is natural to the extent it stays within that region. 
Concretely, given a steering path $\bm{\pi} : [0,1] \to \mathcal{A}$, let $\bm{\gamma}(t) = \bm{p}_{\bm{h} \leftarrow \bm{\pi}(t)}(\bm{x})$ be the behavioral trajectory it induces. 
We define its cumulative output energy:
\begin{equation}
\label{eq:naturalness}
    E_{\text{BC}}(\bm{\gamma}) \;=\; \int_0^1
        d_{\text{BC}}\!\bigl(\bm{\gamma}(t),\, \mathcal{M}_y\bigr)\, dt,
\end{equation}
where $d_{\text{BC}}(\bm{p}, \mathcal{M}_y) = \inf_{\sqrt{\bm{q}} \in \mathcal{M}_y}
d_{\text{BC}}(\bm{p}, \bm{q}) = -\log (\sum_i \sqrt{\bm p_i} \sqrt{\bm q_i})$ is the Bhattacharyya distance to the nearest point on
$\mathcal{M}_y$, a natural choice given that it is simply the negative log of the dot product in Hellinger space (in which $\mathcal{M}_y$ is fit). We note that the formulation above is a tractable proxy 
we use to estimate distance from the model's natural output
distribution, yet it is but one instantiation of a more general
framework we develop in \S\ref{sec:steering-as-metric}.
Applying this measure, we find that manifold steering (weekdays $E_{\text{BC}} = 0.34\pm 0.03$; months $E_{\text{BC}} = 0.36\pm0.01$; letters $2.42\pm0.07$; ages $E_{\text{BC}} = 5.21\pm 0.09$) produces significantly more natural paths, i.e., lower cumulative energy, than linear steering (weekdays $E_{\text{BC}} = 0.93 \pm 0.11$; months $E_{\text{BC}} = 1.09\pm0.06$; letters $6.95\pm0.27$; ages $E_{\text{BC}} = 13.49\pm 0.29$); on average, we see an improvement of a factor of $2.8\times$, with all statistical comparisons yielding $p<0.001$.

We find further verification of the claim above by visualizing output trajectories in behavior space (Fig.~\ref{fig:manifold-steering-main}). Manifold steering consistently traces paths close to $\mathcal{M}_y$, while linear steering cuts through regions far from the behavior manifold, yielding less natural outputs. This result provides causal support for the correspondence between activation and output geometry, and establishes manifold steering as a {principled} form of steering that yields natural behavioral trajectories ($\mathcal{M}_h\rightarrow\mathcal{M}_y$). Next, we explore whether we can find evidence from the opposite direction ($\mathcal{M}_y\rightarrow\mathcal{M}_h$).

\subsection{Behavior Space Geometry Recovers the Activation Manifold}
\label{sec:act2}
\label{sec:results-pullback}

\begin{figure}
    \centering
    \includegraphics[width=\linewidth]{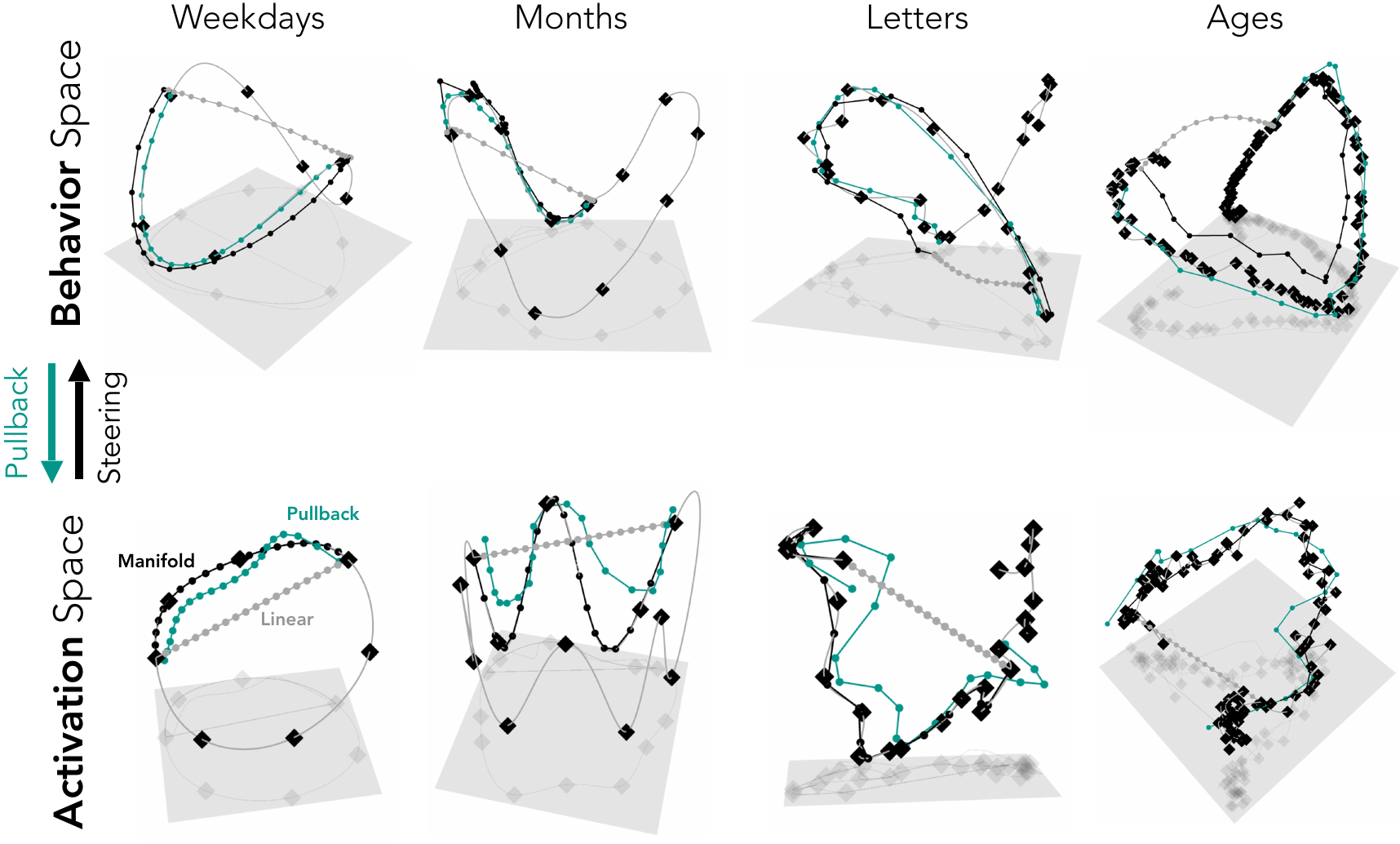}
    \caption{\textbf{Manifold steering and pullback yield coinciding trajectories in activation and behavior space.} Going in the \textbf{Activations$\rightarrow$Behavior} direction, we find that steering along the activation manifold $\mathcal{M}_h$ (black) produces paths that lie close to the behavior manifold $\mathcal{M}_y$. We then examine the reverse direction, \textbf{Activations$\leftarrow$Behavior}: We start with paths along the behavior manifold and optimize for corresponding paths in activation space (i.e., a set of activations that yields the path on $\mathcal{M}_y$ upon intervention). This \textit{pullback} procedure (teal) yields trajectories that resemble the activation manfiold $\mathcal{M}_h$. Thus, we offer bidirectional support for the connection between activation geometry and behavior, and their correspondence reflecting a shared underlying conceptual organization. Paths shown: weekdays `Thursday' to `Sunday'; Months `August' to `December'; Letters `C' to `Q'; Ages 36 to 91.} 
    \label{fig:pullback_main}
\end{figure}

In this section, we aim to uncover whether steering intervention paths optimized to follow the behavior manifold $\mathcal{M}_y$ recover the activation manifold $\mathcal{M}_h$.
To do so, we first take a path $\pi_y^*$ along $\mathcal{M}_y$ in behavior space. We work within the layer and the first 32 dimensions of the subspace in which we fit the activation manifold ($64$ dimensional PCA), and optimize via L-BFGS for a path in activation space which, upon intervention, induces the behavioral path $\pi_y^*$ (see App.~\ref{app:pullback} for further details regarding the optimization procedure). We call the resulting path in activation space the \textit{pullback}.

To quantify how faithfully a pullback path in activation space $\pi_h^{\text{pullback}}$ recapitulates the manifold steering path $\pi_h^*$, we report an \emph{intrinsic} $R^2$. Both paths are projected into a common subspace given by the singular directions explaining $99\%$ of the variance in $\pi_h^*$, restricting the comparison to directions where the path actually extends. We then compute the $R^2$ in this subspace, defining the residual at each point of $\pi_h^{\text{pullback}}$ as its orthogonal closest-point distance to $\pi_h^*$. We compute this score for each optimized pullback path               
$\pi_h^{\text{pullback}}$ and compare with a linear path baseline (see App.~\ref{app:pullback-r2} for more details).

Results are shown in Fig.~\ref{fig:pullback_main}.
We find that the pullback activation paths follow the activation manifold $\mathcal{M}_h$ more closely than the linear steering path, and resemble the shape of the manifold steering paths of \S\ref{sec:act1} (weekdays $R^2_{\text{pullback}} = 0.77\pm 0.03$ vs. $R^2_{\text{linear}} = 0.42\pm 0.07$; months $R^2_{\text{pullback}} = 0.75\pm 0.04$ vs. $R^2_{\text{linear}} = 0.32\pm 0.05$; ages $R^2_{\text{pullback}} = 0.47\pm 0.05$ vs. $R^2_{\text{linear}} = 0.24\pm 0.01$; letters $R^2_{\text{pullback}} = 0.78\pm 0.04$ vs. $R^2_{\text{linear}} = 0.23\pm 0.03$. All statistical comparisons yield $p<0.001$). 
Again we see a striking correspondence between representation and behavior; despite being derived from different sources---manifold
steering from the density of activations and pullback from the
structure of outputs---the two geometries are tightly connected. 

Taken together, these results, alongside those of \S\ref{sec:act1}, provide bidirectional support for the connection between activation geometry and behavior. 
This convergence indicates that $\mathcal{M}_h$ is a core object in the model's representation: the geometry of activation space and the geometry of behavior are alternate views of the same underlying conceptual organization.

\subsection{Unifying Steering Strategies Through Geometry}
\label{sec:steering-as-metric}
In the sections above, we analyzed three methods for steering between two points in activation space that each assume a different geometry: linear steering, which assumes a flat Euclidean geometry; manifold steering, which derives a geometry from naturally occurring activations; and pullback steering, which derives a geometry from naturally occurring output distributions. 
We provided empirical support for our hypothesis that the geometries derived from internal activations and output behaviors are much more similar to each other than the standard Euclidean geometry. 
We now formalize the question of how to steer as \emph{how to choose the right geometry for activation space}.

\paragraph{The Geometry of Steering:} Consider a Riemannian metric
$\bm{G}$, which assigns an inner product at each point of $\mathcal{A}$; together with a path $\bm{\pi} : [0,1] \to \mathcal{A}$, this defines the notion of path length as follows.
\begin{equation}
\label{eq:path-length}
    L_{\bm{G}}(\bm{\pi}) \;=\; \int_0^1
        \sqrt{\dot{\bm{\pi}}(t)^\top\, \bm{G}(\bm{\pi}(t))\,
        \dot{\bm{\pi}}(t)}\, dt.
\end{equation}
Then, a geodesic is defined as the path of minimum length between two endpoints, and each choice of geometry picks out a steering strategy. 
The strategies of linear steering and manifold steering (\S\ref{sec:act1}), written as interpolations in two different coordinate systems (Eqs.~\ref{eq:linear-steering},~\ref{eq:geometric-steering}), are two such choices; the pullback procedure of \S\ref{sec:results-pullback} is a third. 
Now, we make all three geometries explicit.

\begin{definition}[Geometries of Steering]
\label{def:metrics}
Let $E : \mathcal{A} \to \mathbb{R}$ be an energy function such that
$E(\bm{h}) \propto -\log p(\bm{h})$, and let $\bm{g}_y$ be a chosen
Riemannian metric on $\mathcal{M}_y$. We define:
\begin{align}
    \bm{G}_I &= \bm{I}_n,
        &\text{(linear steering)} \\
    \bm{G}_E(\bm{h}) &= \bigl(\alpha\, e^{-E(\bm{h})} + \beta
        \bigr)^{-1} \bm{I}_n,
        &\text{(manifold steering)} \\
    \bm{G}_F(\bm{h}) &= \bm{J}_{\bm{F}}(\bm{h})^\top\,
        \bm{g}_y\bigl(\bm{F}(\bm{h})\bigr)\, \bm{J}_{\bm{F}}(\bm{h})
        + \epsilon\, \bm{I}_n,
        &\text{(pullback)}
\end{align}
where $\alpha, \beta > 0$ are calibration constants, $\epsilon > 0$ regularizes the pullback, $\mathbf{F}:\mathcal{A} \to \behaviorsp$ is the function from naturally occurring activations to naturally occurring behaviors,  and $\bm{g}_y$ is any Riemannian metric on $\mathcal{M}_y$ (e.g., the induced Hellinger metric used in our experiments).
\end{definition}

We discuss the intuitive interpretation of Defn.~\ref{def:metrics} below.

\begin{itemize}[leftmargin=*]

\item \textbf{The Flat Geometry $\bm{G}_I$.}
Linear steering treats activation space as Euclidean: all directions and regions are equally valid, with Geodesics as straight lines $\bm{\ell}(t) = (1{-}t)\,\bm{h}_0 + t\,\bm{h}_1$. This geometry thus encodes no knowledge of naturally occurring activation or outputs.

\item \textbf{The Density Geometry $\bm{G}_E$.}
Manifold steering derives a geometry for activation space from naturally occurring internal representations. Specifically, consider the geometry induced from an energy function $E(\bm{h}) \propto -\log p(\bm{h})$ by rescaling the identity according to local density. Here $e^{-E(\bm{h})}$ plays the role of an unnormalized density: large where activations concentrate (on $\mathcal{M}_h$) and small where they are sparse (off $\mathcal{M}_h$). The inverse makes off-manifold regions expensive and on-manifold movement cheap, with constants $\alpha, \beta > 0$ calibrating the dynamic range~\citep{bethune2025follow}. Geodesics under $\bm{G}_E$ thus follow $\mathcal{M}_h$, recovering manifold steering. 

\item \textbf{The Pullback Geometry $\bm{G}_F$.}
The steering path given by pullback derives geometric structure from naturally occurring model outputs. 
Specifically, $\bm{G}_F$ is the pullback of a chosen geometry on $\mathcal{M}_y$ through the Jacobian of the map from activation space to behavior space $\bm{F} : \mathcal{A} \to \behaviorsp$. 
By construction, path length under $\bm{G}_F$ equals path length of the induced behavioral trajectory along $\mathcal{M}_y$ (up to a regularization term). 
Geodesics under $\bm{G}_F$ are therefore activation paths whose induced behavioral trajectories are geodesics on $\mathcal{M}_y$---exactly the pullback construction of \S\ref{sec:results-pullback}. 
The regularization $\epsilon\,\bm{I}_n$ ensures positive definiteness, since $\bm{J}_{\bm{F}}$ has rank at most $|\mathcal{Z}| - 1 \ll n$; as $\epsilon$ tends to 0, the geometry approaches the pure pullback in the range of $\bm{J}_{\bm{F}}$ and remains Euclidean in its null space.
\end{itemize}

Overall, we claim that while the metrics $\bm{G}_E$ and $\bm{G}_F$ are derived from different sources (internal activations and outputs, respectively), they converge on approximately the same paths in activation space (\S\ref{sec:results-pullback}). 
This suggests the manifolds $\mathcal{M}_h$ and $\mathcal{M}_y$ are two images of the same conceptual geometry, related by an approximate Riemannian isometry.
Consequently, the question of optimally steering model behavior boils down to isolating the geometry of a concept and defining operators to navigate it.

\begin{figure}[!t]
    \centering
    \includegraphics[width=0.95\linewidth]{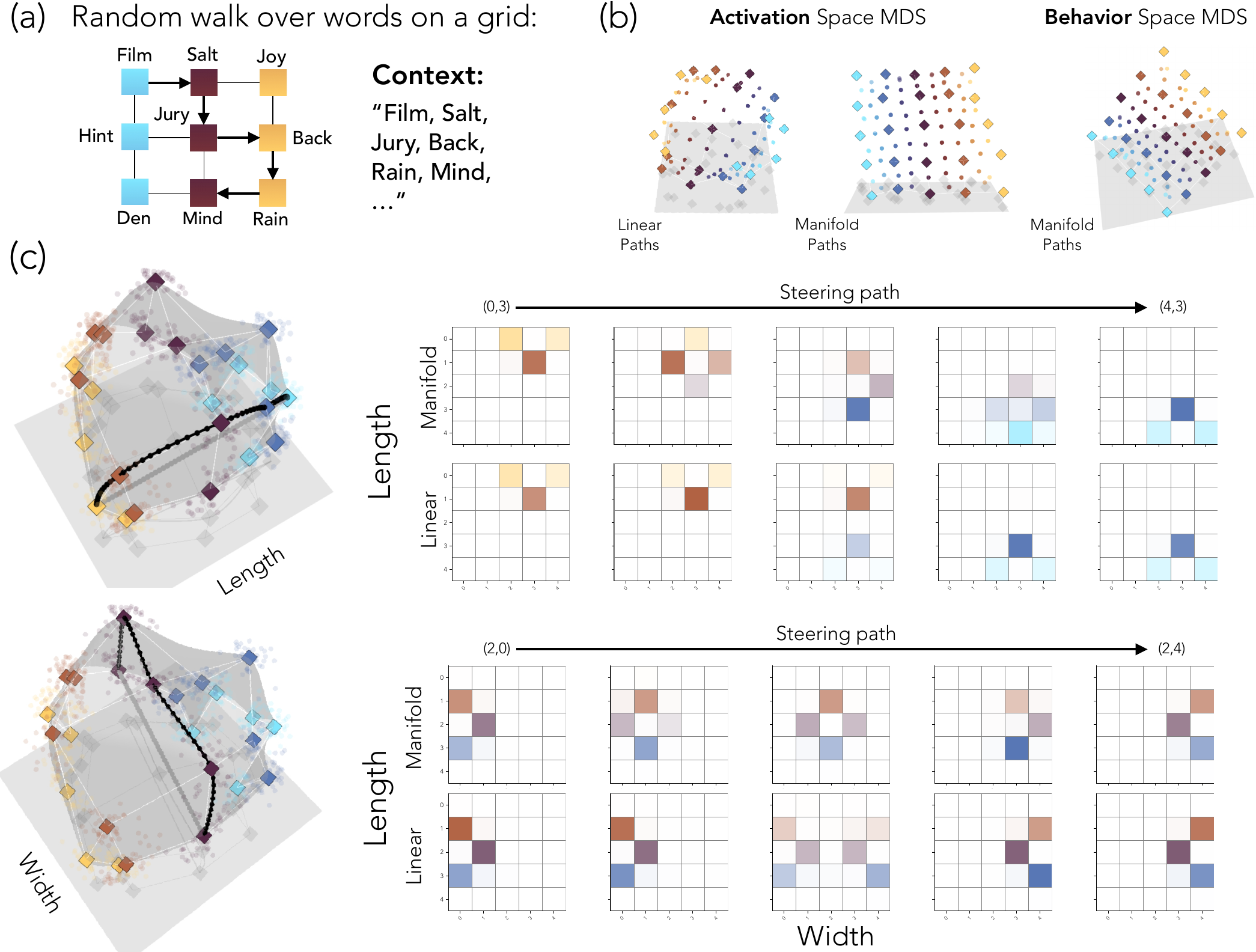}
    \caption{\textbf{Manifold steering enables factored control in multi-dimensional conceptual spaces.} \textbf{(a)} We examine manifold steering on multidimensional spaces using \citet{park2025iclr}'s in-context learning of representations (ICLR) task. In an ICLR task, arbitrary tokens are assigned to nodes along a graph, and a language model is prompted with tokens from a random walk along the graph. \citet{park2025iclr} showed that with sufficient context, models encode the structure of the latent graph in their activations. In this work, we study two graph structures learned in-context ($5\times5$ grid shown above, $9\times9$ cylinder in App. \ref{app:results}). We fit manifolds to activations and output behaviors and intervene on activations using linear and manifold steering. \textbf{(b)} We examine the mapping between the activation and behavior manifolds by computing on-manifold and linear distances in activation space and comparing them to on-manifold distances in behavior space via a multidimensional scaling (MDS) embedding. We find a clear structural match of both the activation and output manifolds with the latent graph, providing direct evidence for these two manifolds encoding a similar underlying conceptual space. In contrast, linear distances in activation space yield a warped structure. 
    \textbf{(c)} We find that manifold steering maintains the quality of smooth and ordered transitions beyond one dimension, and in conceptual spaces learned in-context. Furthermore, we find that steering along one dimension leads to minimal off-target impact, thus affording the appealing quality of factored control. In contrast, linear steering maintains its teleportation behavior.}
    \label{fig:iclr-iclr-grid}
\end{figure}

\section{Manifold Steering Yields Factored Control in Multi-Dimensional Spaces}\label{sec:iclr-iclr}

Our experiments thus far have been limited to one dimensional conceptual spaces arising from training data imbued with real-world structure, i.e., days, months, ages, and letters. In turn, the manifolds we found have been one dimensional curves with a single intrinsic coordinate. Now, we extend our results to a setting with two dimensional conceptual spaces whose geometry are defined via in-context learning. We fit manifolds and show there is a two dimensional intrinsic coordinate system for the manifold, where steering along each coordinate controls an independent dimension of the conceptual space.

\paragraph{In-context learning tasks with synthetic conceptual spaces.}
\citet{park2025iclr} introduce a family of tasks to study the in-context learning of representations (ICLR). For each task, arbitrary tokens are assigned to a discrete graphical structure and language models are supplied with sequences of tokens derived from a random walk on that graph. 
They show that the statistical patterns in the random walk of tokens induce a reorganization of representations that recapitulates the graphical structure used to generate data. This in turn enables the language model to match the next token distribution, i.e., predict tokens adjacent to the current location of the random walk on the grid.
For our two experiments, we assign arbitrary tokens to grid and cylinder graph structures. Thus, the conceptual domain $\mathcal{Z}$ is the set of tokens and the distance metric $d_{\mathcal{Z}}$ is distance on the graph used to generate the random walk. Fig.~\ref{fig:iclr-iclr-grid} (a) shows an example grid and an input prompt generated by a random walk.

\paragraph{Manifold fitting.}
The ICLR grid manifold $\mathcal{M}_h$ is topologically described by a two dimensional surface with no holes or tears. The activation geometry of $\mathcal{M}_h$ is more complex. Its semi-spherical shape shown in Fig.~\ref{fig:iclr-iclr-grid} (c) is induced by task statistics: the random walk visits inner sites more frequently than peripheral sites, leading to slight distortions with respect to the ground truth geometry \citep{park2025iclr, yang2025provable, karkada2026symmetrylanguagestatisticsshapes}. 
We fit two-dimensional sheets to internal activations and output                                            
distributions via thin plate splines (TPS; \citealt{ogsplines, thinplatesplines}), which can be seen as the $2$D analog of the cubic splines used previously. In this case, we use activations corresponding to the last token in the context, and compute centroids according to graph location at a given timestep. Then, TPS finds the smoothest surface interpolating through the centroids (see App.~\ref{app:fit-mh} for further details). 

\paragraph{Isometry results.}
For each ICLR domain, we compute pairwise distance matrices over graph-node centroids under three metrics: Euclidean (linear) distance in the activation subspace, geodesic distance along the fitted activation manifold $\mathcal{M}_h$, and geodesic distance along the behavior manifold $\mathcal{M}_y$. We find very high correlations between geodesic paths on the activation and behavior manifolds ($r=.99$ for both the $5\times5$ grid and  $9\times9$ cylinder domains) and reduced correlations for linear paths ($5\times5$ grid $r=0.90$ ;$9\times9$ cylinder $r=0.81$). To further examine these results, we embed each distance matrix with multidimensional scaling (MDS). Fig.~\ref{fig:iclr-iclr-grid}(b) shows a clean grid structure in the activation manifold and behavior manifold embeddings, while the linear paths in activation space yield a warped surface. 
Again, we see the conceptual space recapitulated in both representation and behavior.

\paragraph{Manifold vs.\ Linear steering.}
We next test whether the fitted two-dimensional activation manifold affords coherent, factored control over the graph geometry used to generate the ICLR inputs.
In particular, we assess whether we can control the position of the random walk input via intervention, and, moreover, whether there is an intrinsic coordinate system where steering along each coordinate independently controls the horizontal and vertical position on the grid.
For each ordered pair of nodes, we use manifold steering and linear steering to interpolate between the start and end centroid and average results over 5 input prompts. 

The top panel of Fig.~\ref{fig:iclr-iclr-grid}(c) shows that manifold steering produces smooth transitions vertically along the steered graph dimension while remaining at the same horizontal position. The bottom panel of Fig.~\ref{fig:iclr-iclr-grid}(c) shows similar smooth transitions but along a horizontal dimension while keeping the same vertical position. This demonstrates the manifold has an intrinsic coordinate system corresponding to the two dimensions of the grid, enabling factored control. Furthermore, this shows the smooth and ordered transitions of manifold steering generalize to multi-dimensional spaces. In contrast, linear steering again fails to provide ordered transitions through grid locations, and shows very clear `teleportation' behavior between the endpoint locations along its path.

\section{Manifold Steering on a Visual World Model: Mountain Car Task}
\label{sec:mountain_car}

We now ask whether the same principles of geometry-aware steering extend to the \textit{visual} domain of world models.
This question is practically motivated: learned world models that predict future observations from past frames and actions are central to model-based reinforcement learning and robotic planning~\citep{ha2018world,hafner2020dreamer,generalist2025gen0, black2024pi0}. If the internal representations of such models admit geometric structure, manifold-based steering could provide a principled mechanism for intervening on a model's behavior through changing its beliefs about the state of the world. 

\begin{figure}[!t]
    \centering
    \includegraphics[width=0.99\linewidth]{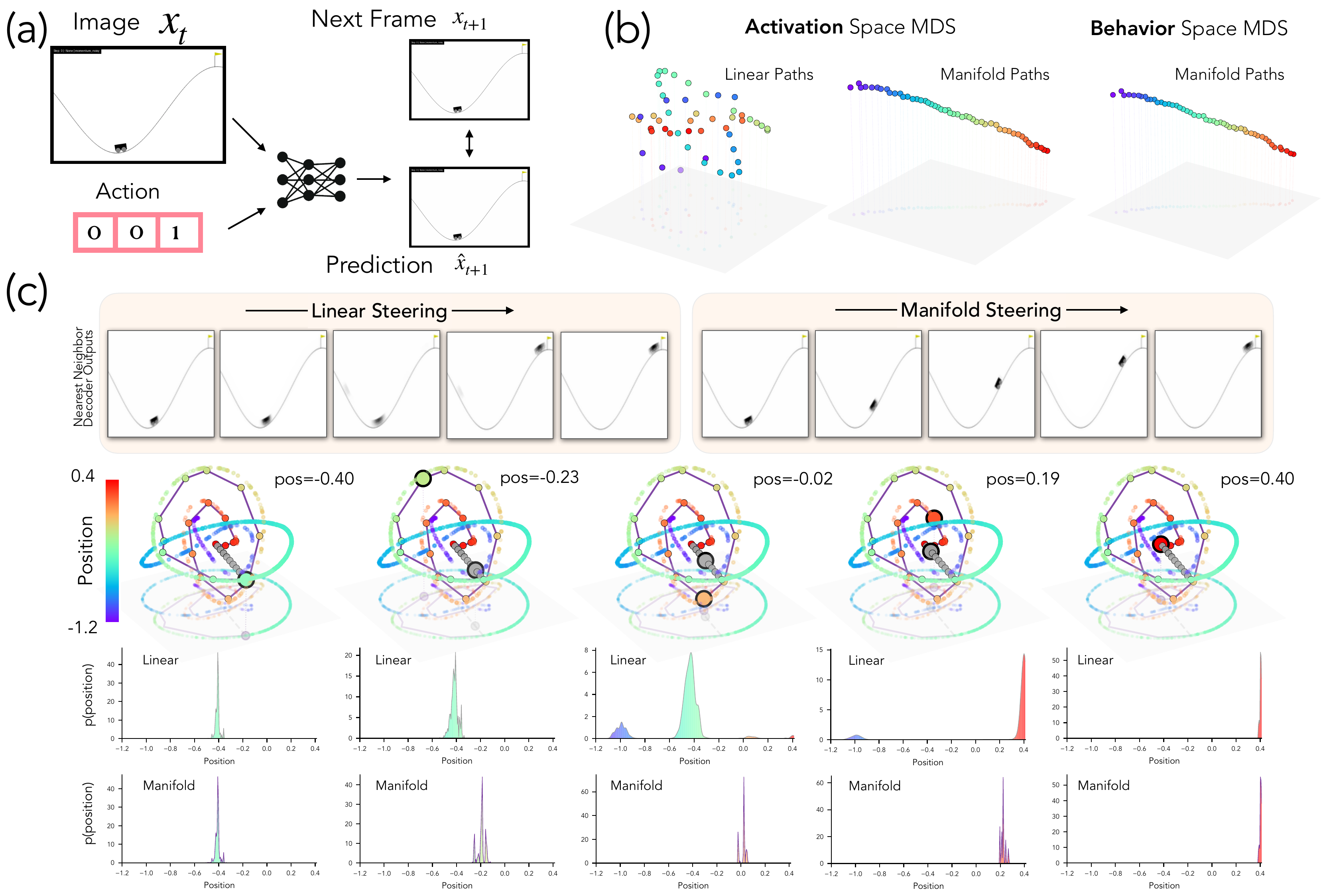}
    \caption{\textbf{Manifold steering on a visual world model produces smooth movement.} \textbf{(a)}. We examine whether manifold steering can generalize to a visual modality by training a recurrent network on the Mountain Car environment \citep{moore1990efficient, sutton2018reinforcement} to predict the next frame $x_{t+1}$ given the previous frame and an action. \textbf{(b)} We test the mapping between the activation and behavior manifolds by computing on-manifold and linear distances in activation space and comparing them to on-manifold distances in behavior space via an MDS embedding. On-manifold paths in activation and behavior space both recover a clean sequential ordering corresponding to location, while the embedding of linear distances scrambles it. \textbf{(c)} \textbf{Middle:} PCA visualization of the activation manifold and five waypoints along a path between $p_A = -0.40$ and $p_B = 0.40$ for both linear (red) and manifold (blue) steering. \textbf{Top}: At intermediate positions along the linear steering path, decoding shows the car as blurred or ambiguously placed, reflecting an incoherent superposition of positional beliefs (\textbf{bottom}) as the path departs from the activation manifold $\mathcal{M}_h$. In contrast, manifold steering along $\mathcal{M}_h$ yields smooth movement of the car up the hill.}
    \label{fig:world_model_steering}
\end{figure}

\paragraph{Environment and model architecture.}
We train a recurrent world model on the Mountain Car environment~\citep{moore1990efficient,sutton2018reinforcement}, a classical control task in which a car must escape a valley by building momentum.
The environment has continuous position $p \in [-1.2, 0.6]$, continuous velocity $v \in [-0.07, 0.07]$, and three discrete actions (left, no-op, right). The model predicts the next frame $x_{t+1}$ given the previous frame $x_t$ and action $a_t$ (see Fig.~\ref{fig:world_model_steering}(a) for an illustration).
The full architecture is shown in Fig.~\ref{fig:mc_architecture}: A convolutional encoder maps each $128 \times 128 \times 3$ RGB frame to a latent vector, $v_t$, which is concatenated with a learned action embedding $e(a_t) \in \mathbb{R}^{16}$ and fed to a Gated Recurrent Unit~(GRU; \citealt{cho2014learning}):
\begin{equation}
    \mathbf{h}_t = \mathrm{GRU}\!\bigl([v_t;\, e(a_t)],\; \mathbf{h}_{t-1}\bigr) \in \mathbb{R}^{n};\;\;
    v_t = \mathrm{LayerNorm}\!\bigl(f_{\mathrm{enc}}(x_t)\bigr) \in \mathbb{R}^{n},
\end{equation}
where $n = 64$. A convolutional decoder produces a residual image from the hidden state, yielding the prediction $\hat{x}_{t+1} = x_t + f_{\mathrm{dec}}(\mathbf{h}_t)$.

\paragraph{Activation and behavior manifold fitting.}
For this setting, we consider \textit{position} to play the role of the conceptual domain $\mathcal{Z} = [p_{\text{min}}, p_{\text{max}}]$ and aim to capture the manifold structure in both activation and behavior space of the vision encoder. To start, we first collect encoder activations from 100 rollouts in the environment (see \S\ref{sec:app_mc_details} for details) and observe they occupy a curved, low-dimensional manifold $\mathcal{M}_h \subset \mathbb{R}^{n}$ (Fig.~\ref{fig:world_model_steering}(c)). 
We parameterize this manifold by partitioning the position range into bins and fit a smooth spline through the means $\{\mu_b\}_{b=1}^{B} \subset \mathbb{R}^{n}$ . For a given input, $x$, we compute the output distribution over positions, $\mathbf{p}(x)$ using the distance of the activations $v(x)$ to the centroid for each position: 
\begin{equation}
\label{eq:mc_belief_operator}
\mathbf{p}(x) \;=\; \mathrm{softmax}\!\left( -\frac{\|v - \mu_b\|_2}{\tau} \right)_{b=1}^{B} \;\in\; \Delta^{B-1},
\end{equation}
with temperature $\tau = 0.5$. We follow \S\ref{sec:naturalness} to parameterize the behavior manifold $\mathcal{M}_y$ by embedding each bin in Hellinger coordinates on the unit sphere and fit a 1D smoothing spline $\gamma_{\mathcal{M}_y} \colon \mathcal{Z} \to \mathbb{R}^{B}$ parameterized by position (full details in App.~\ref{sec:app_mountain_car_details}).
Note that both $\mathcal{M}_h$ and $\mathcal{M}_y$ are 1D structures parameterized by the conceptual coordinate $p$; under PCA visualization (Fig.~\ref{fig:mc_pullback}), both trace closed curves in their respective ambient spaces. The curves are closed because visually distinctive states at the wall, $p \approx -1.2$, and goal, $p \approx 0.4$, are mapped to neighboring activations.

\paragraph{Geometry-aware steering in activation space.}
Fig.~\ref{fig:world_model_steering} compares linear (Eq.~\ref{eq:linear-steering}) and manifold (Eq.~\ref{eq:geometric-steering}) steering through $K=20$ waypoints between encoder states corresponding to positions $p_A = -0.4$ and $p_B = 0.4$, projected into the first three principal components of encoder space.
The geodesic path closely tracks $\mathcal{M}_h$, and the corresponding decoded frames display a smooth, coherent progression of the car through intermediate positions.
The linear path departs from $\mathcal{M}_h$ at intermediate points and decoded frames exhibit blurred or ambiguous car placement, reflecting an incoherent superposition. Then, a `teleportation' to the endpoint is observed, analogous to the behavior of linear steering in the language model experiments.
Moreover, linear steering causes the probability distribution over position to show greater spread compared to on-manifold paths, yielding the ambiguous car placement seen in intermediate points along the path. Finally, we reproduce the pullback procedure in \S\ref{sec:app_mountain_car} and show that optimizing paths in the output distribution over possible car positions results in activation space paths that closely track the $\mathcal{M}_h$ (Fig.~\ref{fig:mc_pullback}).

\paragraph{The geometry assumed by linear steering is not faithful to the conceptual ordering.}
To make the difference between the two steering metrics visually concrete, we apply multidimensional scaling (MDS) to the pairwise distance matrices induced by three different distance functions over $W=50$ anchor positions evenly spaced along $[p_{\text{min}}, p_{\text{max}}]$ (Fig.~\ref{fig:world_model_steering}(b)). Both activation space and behavior space on-manifold distance embeddings recover a clean one-dimensional rainbow ordering of positions, while the linear distance embedding produces a
scrambled three-dimensional structure whose colors are visibly out of order. This is because $\mathcal{M}_h$ folds back on itself in the encoder's ambient space, so two activations whose underlying positions are far apart can sit arbitrarily close in ambient space. Quantitatively, the Pearson correlation between the two arc-length distance matrices is $r=0.99$, while correlation between activation-space linear paths and behavior manifold arc-length falls to $r=0.06$, confirming that the
linear-steering metric is not a faithful proxy for the conceptual ordering that the encoder has learned.

\section{Related Work}
\label{sec:related}

\paragraph{Activation Steering and the Linear Representation Hypothesis.} Activation steering protocols~\citep{bau2018identifyingcontrollingimportantneurons,subramani-etal-2022-extracting,marks2023geometry, panickssery2024steeringllama2contrastive, turner2024steeringlanguagemodelsactivation} are often motivated by the linear representation hypothesis (LRH)---a geometric assumption on model representations~\citep{Smolenksy,elhage2022superposition, park2023linear, costa2025flat, zheng2025model}.
In particular, LRH argues neural networks encode concepts, i.e., latent variables underlying the data distribution~\citep{wang2023concept, rajendran2024causal, rajendran2024learning, okawa2024compositional}, along directions. 
This motivates tools like linear probing~\citep{belinkov2022probing, guerner2023causalprobing} and sparse autoencoders~\citep{cunningham2023sparse, bricken2023monosemanticity, gao2024scaling, bussmann2024batchtopk, fel2025archetypal}.
In the case where the representation geometry for a concept truly aligns with LRH, \cite{bigelow2025beliefdynamicsrevealdual} showed the effects of activation steering on model behavior can be accurately captured by a linear increase in concept log-probability.
However, in the general scenario where geometry of representations does not abide by LRH, the effects of linear steering protocols are less clear.
Recent work has started to fill this gap: e.g., work by \cite{pau2025, linear_adversarial_concept_eraser} has shown that linear steering protocols match the first moments of the current output distribution produced by the model with the target distribution; however, the effects on higher order moments can be unconstrained and adversarial (see results by \cite{sarfati2026shapebeliefsgeometrydynamics}), possibly explaining why it produces incoherent outputs.

In contrast, when the representation geometry is fully respected, our work shows that steering smoothly interpolates the source and the target distributions.
Prior work has shown similar results to this effect in narrow domains, e.g., \cite{engels2024not} ablate representations and write the days-of-the-week circle directly and \cite{kantamnenihelix} follow a similar protocol for a helix representing numbers, but they lack a more general account of how representation geometry and output behavior map on to each other.
The closest work to ours is the contemporary paper by \cite{park2026information}, who study a toy model where the representation-to-output distribution mapping is described via a simple softmax operation.

\paragraph{Activation Geometry and its Origins.} A large body of recent work has shown neural networks encode concepts along nonlinear, curved geometries embedded in low-dimensional subspaces across both modalities and architectures~\citep{fel2025into, pearce2025tree, yocum2025neural, model2025, lubana2025priors, costa2025flat, park2025iclr, engels2024not, karkada2026symmetrylanguagestatisticsshapes, shai2024transformers, shai2026transformerslearnfactoredrepresentations, saxe2019mathematical, park2024geometry, morwani2024feature, kantamnenihelix, song2023uncovering, zhou2025fone, maheswaranathan2019universality}.
While earlier work \citep{saxe2019mathematical, arora-etal-2018-linear, park2023linear, yocum2025neural} concretized, in toy settings, how structure in the data-generating process imposes geometric constraints on a neural network's representations, only recently have such accounts been extended to make predictions about the geometry of neural representations at scale (e.g., \cite{merullo2025linearrepresentationspretrainingdata}).
\cite{karkada2026symmetrylanguagestatisticsshapes} and \cite{ korchinski2025emergence} argue that symmetries in data statistics enforce geometries best suited for reflecting the uncertainty of the distribution in a model's representation  (cf. \citealt{prieto2026correlations}), and offer plausible accounts for the formation of representations in-context, as shown by works such as \cite{park2025} and \cite{lepori2026language}.

\paragraph{Causal Analysis of Neural Networks.} Several works have convincingly argued that tools like probing or visualization of representations are insufficient to make claims about model behavior, i.e., artifacts produced via these tools can yield misleading explanations for why a model behaves the way it does~\citep{geiger-etal-2020-neural, belinkov2022probing, bolukbasi2021interpretability, saphra2024}.
As such, a vast array of research has used intervention on activations to study model internals \citep{li2017understandingneuralnetworksrepresentation, giulianelli-etal-2018-hood, cammarata2020thread:, elazar-etal-2020, linear_adversarial_concept_eraser, ravfogel2023, kernelized_concept_Eraser, leace, geva2023dissectingrecallfactualassociations, meng2022locating,meng23, vig2020, geiger-etal-2020-neural, Davies:2023, DBLP:conf/emnlp/StolfoBS23, guerner2023causalprobing,wang2023interpretability,todd2024,arora2024causalgym, huang-etal-2024-ravel, feng2024how, mueller2025mibmechanisticinterpretabilitybenchmark,prakash2025languagemodelsuselookbacks, gur-arieh-etal-2025-enhancing, grant2025emergentsymbollikenumbervariables, diego2024}.
This interpretability research leverages the frameworks of causal mediation \citep{pearl2001directindirecteffects, vig2020, mueller2026} and causal abstraction~\citep{rubinstein2017,beckershalpern2019,geiger2021, geiger2025ca,geiger2026} to ground understanding of model internals in the theory of causality~\citep{hume1748enquiry, pearl1999, Spirtes}.

\section{Discussion}

\paragraph{Geometry-aware steering reveals the shared structure of behavior and representation.} We build out an empirical phenomenology that relates structure in activation space to the model output behavior. First, we show an isometry between representations and behavior manifolds, i.e., distance between two points on the activation manifold $\mathcal{M}_h$ aligns with distance between the distributions induced by those points on the behavior manifold $\mathcal{M}_y$. Second, we show that steering representations along geodesics on $\mathcal{M}_h$ induces smooth, coherent transitions in behavior that follow geodesics on $\mathcal{M}_{y}$. Third, we show that optimizing interventions to produce behaviors following geodesics on $\mathcal{M}_y$ recover trajectories in activation space that follow $\mathcal{M}_h$. Thus, we establish a causal bridge between representation and behavior that reveals shared structure reflecting underlying conceptual geometry.

Our results also suggest that pathologies of linear steering---brittleness, incoherence, off-target effects \citep{wu2025axbenchsteeringllmssimple, bigelow2025beliefdynamicsrevealdual, da2025steering, bhalla2024towards, tan2024analysing}---stem from the mismatch between assumed flat geometry and the true curved geometry of representation space, rather than an inherent challenge with representation-based intervention. 
This reframes the challenge of steering from ``finding the right direction'' to ``finding the right geometry''.

\paragraph{Where does the shared geometry of behavior and representation come from?} 
While we do not study the origins of the shared geometry between behavior and representation, our experimental results are consistent with the hypothesis that conceptual structure constrains the geometry of both representation and behavior. 
While data statistics shape the geometry of neural representations \citep{merullo2025linearrepresentationspretrainingdata, karkada2026symmetrylanguagestatisticsshapes, prieto2026correlations}, this fails to explain how geometric structure is formed for out-of-distribution inputs.
For example, our in-context learning tasks (Sec.~\ref{sec:iclr-iclr}) have synthetically defined geometries that imbue tokens with contextual meaning that is wildly different from the meaning learned during training. 
As such, the model must form novel representations and produce novel behaviors.
The fact that we are able to establish a shared geometry for representation and behavior in these novel in-context learning tasks suggests that regardless of how training data statistics inform the geometries seen in the model, the output behavior is now computationally constrained by the activation geometry (see the contemporary work by \cite{yocum2025neural} for a formalization of this claim).

\paragraph{Intrinsic coordinates of representation manifolds as units of causal analysis.}
\cite{mueller2024} frames the field of mechanistic interpretability as being on a quest to discover a primitive unit of representation best suited for the causal analysis of neural network internals.
Causal abstraction provides a theoretical framework for defining such units of analysis \citep{geiger2025ca, geiger2026}, however \cite{sutter2025nonlinearrepresentationdilemmacausal} point out that allowing arbitrarily complex units admits degenerate solutions.
Our work suggests a path toward both answering \cite{mueller2024} and addressing the problem identified by \cite{sutter2025nonlinearrepresentationdilemmacausal}: the appropriate units of causal analysis are intrinsic coordinates on manifolds in activation space, and fitting these manifolds to naturally occurring activations provides a constraint that helps rule out degenerate solutions (cf. \citealt{grant2026addressingdivergentrepresentationscausal}).

\section{Future Work and Limitations} 

The goal of our paper was to understand the role of geometry in neural networks and, subsequently, use this understanding to concretize what it means to steer model behavior via representations.
We have shown that the geometry of neural network representations provides a blueprint for effective control. When interventions respect the geometry of activation space, the change in behavior is smooth and coherent; when they ignore it, they risk producing states with no natural behavioral counterpart. 
While we believe our results have enabled significant progress towards the motivating goals, there are remaining limitations that need to be addressed in future work.

\begin{itemize}[leftmargin=*]
    \item \textbf{Expanding experimental validation to more complex domains.} To illustrate our arguments, we focused on simple settings for which the concept of interest had a well-defined domain (e.g., weekdays), and the expected task outputs are the concepts themselves, therefore the conceptual geometry is directly displayed in the outputs.
    To further validate the claims posited in this paper, future work is needed to explore more abstract concepts, e.g., refusals~\citep{arditi2024refusal}, sycophancy~\citep{vennemeyer2025sycophancy}, and persuasion~\citep{costello2026large}.
    For such concepts, output behavior will likely reflect conceptual structure in subtler ways, and it remains to be tested whether the conceptual geometry of such concepts can be inferred from behavior and related to representations as we did in this work.  
    Moreover, in these more complex cases, it is unclear what are the right primitives for a representational account; we may need a notion of dynamics over manifolds, a view of representation as an aggregation of several geometric structures (similar to results seen by \cite{fel2025into} in a vision context), or perhaps an altogether different object.
    Even if geometry is the right substrate to work with, we emphasize the simplicity of our domains allowed us to easily isolate the target concept's geometry via synthetic, template-based text.
    Moving to more complex scenarios will require isolating the geometry of concepts from in-the-wild data.

    \item \textbf{Moving from token to sequence-level outputs.} Another way in which our tasks are simplified is our focus on the next-token distribution.
    This makes analysis feasible and helps avoid the combinatorial complexity involved in studying multi-token sequences.
    The obvious way to expand from our work's token-level focus to sequence-level focus involves formalizing arguments in the language of ``beliefs'', i.e., latent variables underlying the posterior predictive induced by a model in response to an input~\citep{bigelow2023context, bigelow2025beliefdynamicsrevealdual, wurgaft2025incontextlearningstrategiesemerge}.
    Correspondingly, what we expect to see via geometry-aware interventions is the nature of output sequences produced by a model will change as we perform steering: e.g., navigating the geometry of sycophancy (were it to exist) should allow us to alter the extent or type of sycophancy exhibited in model outputs; however, this property will be latent, rather than a concrete token-level change.
    
    \item \textbf{Fitting the geometry.} While we used a specific protocol to fit the observed geometries~\citep{thinplatesplines}, we note there is a rich literature on fitting low-dimensional manifolds~\citep{coifman2006diffusion, brand2002charting, scholkopf1997kernel, roweis2000nonlinear, jones2024diffusion, jones2026computing,meila2023}.
    Critically, beyond just fitting the manifold, what we seek is an operator that allows us to navigate the manifold. 
    For the domains analyzed in this work, we have ground-truth knowledge about how different states of the concept relate to each other, which allows us to define intrinsic coordinates for spline fitting.
    An unsupervised protocol would however significantly broaden the applicability of our methods. 

    \item \textbf{Manipulating intermediate algorithmic variables.}
    All of our experiments are about manipulating the output behavior of neural networks directly. 
    However, the most interesting control protocols will require manipulating intermediate quantities that mediate the flow of information from input to output, e.g., an image model determining the shape of an object in service of predicting its weight. 
    
\end{itemize}

\section*{Acknowledgments}

The authors thank David Klindt, David Bau, Thomas Icard, Jing Huang, and the Mechanisms team at Goodfire for helpful conversations during the course of this project.

\newpage
\bibliographystyle{colm2026_conference}
\bibliography{references}

\newpage
\appendix

\section{Experimental Details for Language Tasks}\label{app:experimental-details_language}

This Appendix describes the procedures behind the experiments of Sections \S\ref{sec:geometry}, \S\ref{sec:phenomenology}, and \S\ref{sec:iclr-iclr}. The following section (Appendix \S\ref{sec:app_mountain_car_details}) will provide details for the mountain-car experiment in  \S\ref{sec:mountain_car}.

\subsection{Tasks and Datasets}

\paragraph{Natural domain tasks.}
We use four natural-domain addition tasks: weekdays and months (cyclic), and letters and ages (sequential). The full templates and entity sets are listed in Table~\ref{tab:natural-domains}. For each task, we enumerate every (entity, increment) pair whose result lies in the task's target set, dropping pairs whose result would fall outside it (e.g.\ letters past Z, or ages outside $[10, 100]$). The reported activations and output distributions are computed at the answer-token position, and concept centroids are obtained by averaging across all prompts whose ground-truth result is the same value of $\mathcal{Z}$.

\begin{table}[h]
\centering
\small
\begin{tabular}{@{}l p{5.2cm} p{1.9cm} p{1.6cm} l r@{}}
\toprule
\textbf{Task} & \textbf{Template} & \textbf{Entities} & \textbf{Increments} & \textbf{Structure} & \textbf{$|\mathcal{D}|$} \\
\midrule
Weekdays &
Q: What day is \{k\} days after \{entity\}?\textbackslash nA: &
\textit{Monday}, \dots, \textit{Sunday} (7) &
\textit{one}, \dots, \textit{seven} &
cyclic & 49 \\
\midrule
Months &
Q: What month is \{k\} months after \{entity\}?\textbackslash nA: &
\textit{January}, \dots, \textit{December} (12) &
\textit{one}, \dots, \textit{seven} &
cyclic & 84 \\
\midrule
Letters &
Consider letters in the alphabet. Starting at letter \{entity\}, we increment by \{k\}. The result is letter &
\textit{C}, \dots, \textit{Z} (24) &
\textit{one}, \textit{two} &
sequential & 48 \\
\midrule
Ages &
Alice is \{entity\} years old. Bob is \{k\} years older than Alice. Q: How old is Bob?\textbackslash nA: Bob is &
\textit{1}, \dots, \textit{99} &
\textit{1}, \dots, \textit{10} &
sequential & 909 \\
\bottomrule
\end{tabular}
\vspace{0.5em}
\caption{Natural-domain arithmetic tasks. For cyclic domains, results wrap around the modulus (7 for weekdays, 12 for months); for sequential domains, (entity, increment) pairs whose result falls outside the target set are filtered. The last column reports the dataset size $|\mathcal{D}|$ after this filter.}
\label{tab:natural-domains}
\end{table}

\paragraph{In-context learning of representations.}
For the multi-dimensional setting of \S\ref{sec:iclr-iclr}, we use the in-context learning of representations (ICLR) family of \citet{park2025iclr}: arbitrary tokens corresponding to nouns (e.g., "film", "rain") are assigned to the nodes of a graph, and prompts are random walks on that graph. We study a $5 \times 5$ grid and a $9 \times 9$ cylinder, with random walks of $2048$ entity tokens. As in \citet{park2025iclr}'s setup, the random walks we sample do not allow backtracking, which we find aids models in learning the underlying structure. 

\subsection{Model, Intervention Site, and Output Distribution}\label{app:model-and-site}

We investigate Llama 3.1 8B \citep{touvron2023llama} activations at layer 28 in bfloat16 for all tasks. All interventions are performed on the residual stream at the last-token position. We chose to examine a late layer of the model to ensure that concept geometries are fully computed.

For an input $x$, the output distribution $\bm{p}(x) \in \mathcal{Y}$ used throughout the paper is constructed as follows: we softmax over the full vocabulary logit distribution and aggregate probability mass over each concept value's variant token spellings (e.g.\ the tokens \texttt{` Monday'}, \texttt{`Monday'}, and \texttt{`monday'} are all summed into the \texttt{Monday} entry). The remaining probability mass on tokens not associated with any concept value is collected into a single \emph{`other'} bin, yielding a distribution on the open simplex $\Delta^{|\mathcal{Z}|}$ over $|\mathcal{Z}|+1$ classes.

\subsection{Fitting the Activation Manifold $\mathcal{M}_h$}\label{app:fit-mh}

To identify the activation manifold $\mathcal{M}_h$, we first obtain points in full activation space and transform them into a 64-dimensional subspace obtained via PCA over the activations $\bm{h}(x)$ across all prompts in the task. The manifold lives entirely in the 64-dimensional PCA subspace; the orthogonal complement is preserved during all subsequent interventions (\S\ref{app:steering}).

We compute concept centroids $c_i$ as the mean of the projected activations across all prompts whose ground-truth result equals the $i$-th concept value, and fit a smooth interpolant through them. For the four natural-domain tasks the interpolant is a one-dimensional cubic spline \citep{reinsch1967smoothing}: a natural cubic spline (with vanishing second derivatives at the endpoints) for the sequential tasks (letters, ages), and a periodic cubic spline for the cyclic tasks (weekdays, months) so that the curve closes smoothly. For the sequential tasks we use the ground-truth ordinal index of each concept as its intrinsic coordinate. For the cyclic tasks the centroids form a near-circular loop in the top two principal components of the activation subspace, so we instead derive the intrinsic coordinate $\theta = \operatorname{atan2}(\mathrm{PC}_2, \mathrm{PC}_1)$ in an unsupervised manner.

The interpolant for the ICLR tasks is a thin-plate spline (TPS; \citet{ogsplines, thinplatesplines}), a multi-dimensional generalisation of the cubic spline which minimizes the bending energy $\int \|\nabla^2 f\|^2$. Thin-plate splines map points in a lower-dimensional intrinsic space to the full ambient space. The TPS parameterisation requires a choice of intrinsic coordinates for the centroids. We use the ground-truth graph coordinates of each node in the ICLR task as intrinsic coordinates. Both the grid and cylinder tasks use the 
standard TPS kernel $r^2 \log r$; for the cylinder, which has both a linear and a periodic dimension, we additionally apply a ghost-point procedure where each control point is duplicated at one period above and below its $\theta$ value (and we drop the linear-in-$\theta$ polynomial column) to enforce closure across the periodic dimension. In every case the spline interpolates the centroids exactly, so $\mathcal{M}_h$ passes through every $c_i$.

\subsection{Fitting the Behavior Manifold $\mathcal{M}_y$}\label{app:fit-my}

Behavior centroids $b_i = \bar{\bm{p}}_i$ are computed analogously to the activation centroids, by averaging the model's output distributions across all prompts whose ground-truth result equals the $i$-th concept value. Because the probability simplex is not a proper metric space, we map each centroid into Hellinger coordinates via $b_i \mapsto \sqrt{b_i}$, placing it on the non-negative orthant of the unit $\ell_2$ sphere in $\mathbb{R}^{|\mathcal{Z}|+1}$. We then fit the same family of splines used for $\mathcal{M}_h$ to the Hellinger-embedded centroids: a 1D cubic spline (natural or periodic) for the natural-domain tasks, and a thin-plate spline for the ICLR tasks. Analogous to the activation manifold, the fit passes exactly through every $\sqrt{b_i}$ and we do not apply a smoothing penalty.

The spline is fit in Euclidean space, but valid $\sqrt{b_i}$ points lie on a curved sphere. A naive fit to their ambient coordinates would leave the sphere between centroids, and an off-sphere vector does not square to a valid distribution. We therefore fit the spline in the \emph{tangent plane} of the sphere at a base point $b_*$ -- a flat space that touches the sphere at $b_*$ -- and lift back to the sphere at decode time.

We take $b_*$ to be the Euclidean mean of $\{\sqrt{b_i}\}$, re-normalized to unit length; because every $\sqrt{b_i}$ lies in the non-negative orthant, so does $b_*$. The \emph{log-map} $t_i = \log_{b_*}\!(\sqrt{b_i})$ projects each centroid onto the tangent plane: it returns a vector whose direction points from $b_*$ along the geodesic to $\sqrt{b_i}$ and whose length equals that geodesic distance. We fit the spline to the tangent vectors $\{t_i\}$. To decode at a query coordinate $u$, we evaluate the spline to obtain a tangent vector $t$ and apply the \emph{exponential map} $\exp_{b_*}(t)$, the inverse of the log-map: it walks distance $\|t\|$ along the geodesic on the sphere starting at $b_*$ in direction $t$. The result is unit-norm by construction, so $\mathcal{M}_y$ stays on the sphere everywhere, and because $\exp_{b_*}\!\circ\log_{b_*}$ is the identity on the sphere, the decoded curve passes through every $\sqrt{b_i}$ exactly.

\subsection{Geodesic Distances and the Isometry Test}\label{app:geodesics}

To compute the geodesic distance $d_{\mathcal{M}_h}(c_i, c_j)$ between two concept centroids, we discretize the line segment between $\bm{s}^{-1}(c_i)$ and $\bm{s}^{-1}(c_j)$ in intrinsic coordinates into 150 equal sub-intervals, decode each waypoint through $\bm{s}$, and accumulate consecutive ambient distances. Each waypoint therefore lies on $\mathcal{M}_h$ by construction, and the resulting arc length is measured in the 64-dimensional PCA subspace in which the manifold lives, with the Euclidean norm as the ambient norm. For the behavior manifold $\mathcal{M}_y$ we follow the same procedure but compute distances in the full sqrt-probability ambient space $\mathbb{R}^{|\mathcal{Z}|+1}$ rather than a PCA-reduced subspace, using the Hellinger distance $d_H(p, q) = \tfrac{1}{\sqrt{2}}\|\sqrt{p} - \sqrt{q}\|_2$ directly on the sqrt-embedded waypoints.

The isometry score reported in \S\ref{sec:geometry} is the Pearson correlation between the upper-triangular entries of the resulting pairwise distance matrices. We augment the $W$ centroid vertices with $K$ interior points sampled at equally spaced fractions of the u-space geodesic between each centroid pair, decoded onto the manifold via $\bm{s}$ so that every vertex lives on $\mathcal{M}_h$ or $\mathcal{M}_y$. We choose $K$ so that the vertex set is dense enough to probe the geometry between centroids: $K = 4$ for weekdays ($W = 7$); $K = 1$ for months ($W = 12$), alphabet ($W = 24$, letters $C\text{--}Z$), and the grid $5{\times}5$ task ($W = 25$); and $K = 0$ for age ($W = 91$, ages $10\text{--}100$) and the cylinder $9{\times}9$ task ($W = 81$), whose centroids are already dense. We then correlate every off-diagonal pair in the full vertex set except those whose two vertices lie on a common centroid-pair geodesic, since those distances are sub-arcs of the same geodesic and would inflate the correlation by construction. To visualise the resulting pairwise structure, we embed each distance matrix into three dimensions via classical multidimensional scaling, as shown in Figs.~\ref{fig:shapes-wrapfigabove_below}, ~\ref{fig:shapes-wrapfigabove_below_repeat}, and~\ref{fig:iclr-iclr-grid}.

\subsection{Steering Interventions}\label{app:steering}

For each pair of concept values $(z_a, z_b)$, we steer the model from the centroid $c_a$ to the centroid $c_b$ via a path of $K = 50$ waypoints. We use a fixed set of base prompts sampled randomly from the task's input distribution (the prompts' ground-truth results vary, and the same set is reused across all pairs): $16$ prompts for the natural-domain tasks and $5$ for the ICLR tasks. At each waypoint $\bm{\pi}(t)$, we intervene at the last-token residual-stream activation of the target layer and continue the forward pass to obtain $\bm{p}_{\bm{h} \leftarrow \bm{\pi}(t)}(x)$. Every reported behavioral trajectory is the pointwise mean over the base prompts. We use up to 50 randomly-sampled pairs per task. On the smaller-domain tasks where $W \cdot (W-1) < 50$, all pairs are used.

The two steering strategies differ both in how the waypoints $\bm{\pi}(t)$ are constructed and in what the intervention replaces (Eqs.~\ref{eq:linear-steering}, \ref{eq:geometric-steering}). For \emph{manifold} steering, $c_a, c_b$ live in intrinsic coordinates and the path is the manifold geodesic between them; at each waypoint we decode $\bm{\pi}(t)$ onto $\mathcal{M}_h$ in the 64-dimensional PCA subspace, lift it back to the residual-stream basis via the PCA inverse, and combine it with the prompt's unchanged off-subspace residual -- so the steered activation differs from the base only in its top-64 PCA components. For \emph{linear} steering, $c_a, c_b$ are the raw activation centroids in the full residual stream, the path is the straight line between them, and the entire residual-stream activation is replaced by $\bm{\pi}(t)$ at each step.

\subsection{Naturalness Metric}\label{app:naturalness}

The cumulative output energy $E_{\text{BC}}$ of \S\ref{sec:naturalness} (Eq.~\ref{eq:naturalness}) is computed as the sum of the Bhattacharyya distances $D_{\text{BC}}(\bm{\gamma}(t), \mathcal{M}_y) = -\log \sum_i \sqrt{\gamma_i(t)\, q_i(t)}$ between the induced output distribution at each of the $K=50$ waypoints along steering paths and the closest point $q(t)$ on $\mathcal{M}_y$. We use the Bhattacharyya distance because $\mathcal{M}_y$ is fit in Hellinger geometry and the two are tightly related, $D_{\text{BC}} = -\log(1 - d_H^2)$, so $D_{\text{BC}}$ stays inside the same geometry the manifold was constructed in. For each of the up to 50 sampled centroid pairs, we average the per-waypoint cumulative sum across the base prompts to obtain one scalar per pair, and report the mean and standard error of these per-pair scalars.

\subsection{Pullback Optimization}\label{app:pullback}

The pullback procedure of \S\ref{sec:results-pullback} consists of two stages. First, we fix a behavioral target by evaluating the spline geodesic on $\mathcal{M}_y$ between the two behavior centroids $b_a$ and $b_b$ at $K = 20$ uniform fractions, yielding a sequence of target distributions $\hat{\bm{p}}_t \in \mathcal{M}_y$. Second, we optimize an activation-space path $\pi_h^{\text{pullback}}$ which, when used to intervene at each waypoint, induces a behavioral trajectory matching $\hat{\bm{p}}_{0:K}$.

\paragraph{Path Parameterization.}
We parameterise $\pi_h^{\text{pullback}}$ as a one-dimensional natural cubic spline through 10 control vectors at uniform $t$-positions, all of which are optimisation variables. The path is evaluated at the same $K = 20$ uniform fractions used to generate the target. Each control vector is restricted to the first 32 PCA components of the 64-dimensional subspace; the remaining 32 components, together with the orthogonal residual, are held at the base prompt's activation values during the intervention. We note that the linear and manifold-steering paths used as comparisons span the full 64-dimensional subspace, so the pullback optimization is operating within a strictly smaller search space.

\paragraph{Loss and optimizer.}
The loss at each waypoint $t$ is the squared Hellinger distance $d_H^2(\bm{p}_{\bm{h} \leftarrow \bm{\pi}(t)}(x_n), \hat{\bm{p}}_t)$ between the induced output distribution and the target, averaged over 16 base prompts $\{x_n\}$ sampled freshly per pair, each conditioned on ground-truth $z_a$ (in contrast to the steering setup of \S\ref{app:steering}, where a fixed unfiltered set is reused across pairs). We minimize the sum of these per-waypoint losses with L-BFGS using strong-Wolfe line search, running 50 outer steps with up to 5 inner iterations each. The optimization is initialized by linearly interpolating between the two centroids in the 32-dimensional subspace and then sampling the resulting line at the 10 control-$t$ positions. We stop early when the relative change in loss between two consecutive outer steps falls below $10^{-3}$. We disable the path-norm regularizer for weekdays; on the other three natural-domain tasks we add a small regularizer that penalizes deviations of $\|\bm{\pi}(t)\|$ from the linear interpolation between the endpoint centroid norms $|c_a|, |c_b|$. We use weight $10^{-3}$ for age and $5 \times 10^{-4}$ for months and alphabet. This discourages the optimizer from drifting into a high-norm shortcut basin.

\subsection{Pullback Recovery $R^2$}\label{app:pullback-r2}

To compare the optimised pullback path $\pi_h^{\text{pullback}}$ to the manifold-steering path $\pi_h^*$ along $\mathcal{M}_h$, we project both into the SVD basis of $\pi_h^*$ that captures at least $99\%$ of $\pi_h^*$'s variance. In this basis we define the residual at each pullback waypoint as its orthogonal closest-point distance to $\pi_h^*$, and report
\begin{equation*}
R^2 \;=\; 1 - \frac{\sum_t \|\pi_h^{\text{pullback}}(t) - \mathrm{proj}_{\pi_h^*}\pi_h^{\text{pullback}}(t)\|^2}{\sum_t \|\pi_h^{\text{pullback}}(t) - \bar{\pi}_h^{\text{pullback}}\|^2}.
\end{equation*}
The linear baseline used in the same comparison is the straight chord between $c_a$ and $c_b$ in the 64-dimensional PCA subspace---not the linear-steering trajectory after intervention. As in \S\ref{app:naturalness}, the values reported in \S\ref{sec:results-pullback} are mean $\pm$ standard error across the per-pair scalars, with $p$-values from paired $t$-tests against the linear baseline.

\newpage
\section{Experimental Details for the Vision Task}
\label{sec:app_mountain_car_details}

This section contains additional details on the experiment from \S\ref{sec:mountain_car}.

\subsection{Mountain Car.}\label{sec:app_mc_details}
\paragraph{Data collection.}
To recover the encoder's position manifold we harvest activations on $100$ rollouts collected in \texttt{MountainCar-v0} (max $200$ steps per episode) under a mixed stochastic policy chosen to give broad coverage of the position-velocity state space.  At the start of each episode we sample one of two policies: with probability $0.7$, a \emph{noisy momentum} policy that pushes in the direction of the current velocity but, at each step, replaces the action with a uniform random action with probability $0.4$; with probability $0.3$, an \emph{oscillating square-wave} policy that alternates between full-left and full-right thrust on a fixed period sampled uniformly from $\{5,\dots,25\}$ steps.  We then pass each rendered frame through the trained encoder, label the resulting activation with the underlying ground-truth position, and fit the manifold to this collection of position-labelled activations.

\begin{figure}[!t]
    \centering
    \includegraphics[width=0.99\linewidth]{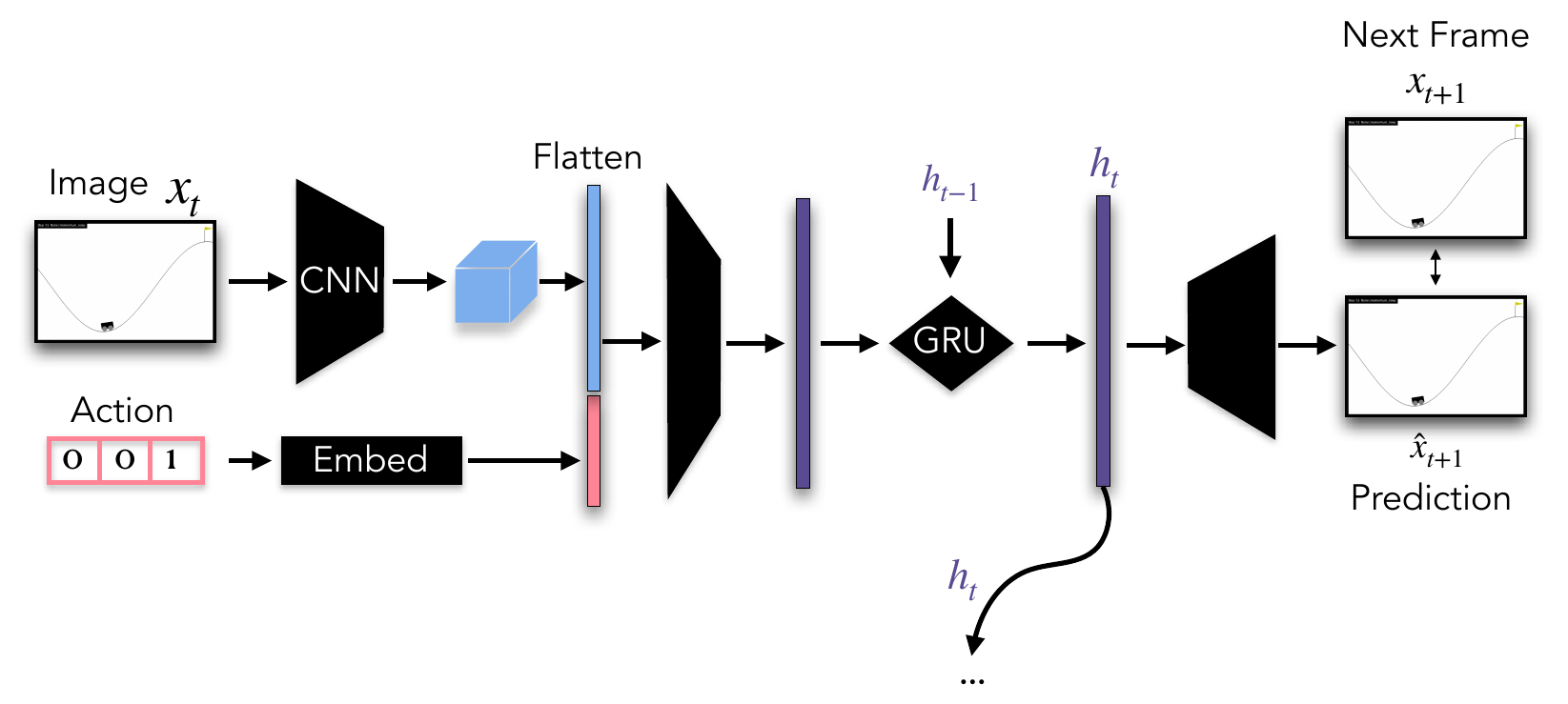}
    \caption{\textbf{Recurrent visual world-model architecture.}
    A convolutional encoder $f_{\mathrm{enc}}$ maps each
    $128\times 128\times 3$ frame $x_t$ to a layer-normalized latent $z_t \in \mathbb{R}^{n}$ with $n=64$. The discrete action $a_t \in \{0, 1, 2\}$ is mapped to a learned embedding
    $e(a_t) \in \mathbb{R}^{16}$, concatenated with $z_t$, and fed to a GRU together with the previous hidden state $\mathbf{h}_{t-1}$. A convolutional decoder $f_{\mathrm{dec}}$ produces a residual image from the resulting hidden state $\mathbf{h}_t$, yielding the next-frame prediction $\hat{x}_{t+1} = x_t + f_{\mathrm{dec}}(\mathbf{h}_t)$, supervised against the ground-truth frame $x_{t+1}$.}
    \label{fig:mc_architecture}
\end{figure}

\paragraph{Manifold fitting.}
To parameterize the activation manifold, $\mathcal{M}$, we partition the position range into $B=100$ bins, compute the mean encoder output per occupied bin, and fit a smoothing spline $\gamma_{\mathcal{M}} \colon [0,1] \to \mathcal{M}$ through these means (one univariate spline per coordinate, weighted by the square root of bin counts to regularize sparse regions). We additionally verify via linear probing that the encoder representations $z_t$ encode the ground-truth physics: a Ridge regression probe recovers position with $R^2 \approx 0.95$ and velocity with $R^2 \approx 0.90$.
A three-component PCA of the encoder outputs reveals the spline $\gamma_{\mathcal{M}}$ as a curve that closely tracks the data manifold, while the chord $\ell$ between the same endpoints cuts through its interior.

To parameterize the behavior manifold, $\mathcal{M}_y$,  discretize $\mathcal{Z}$ into $B$ bins with centers $\{\mu_b\}_{b=1}^{B} \subset \mathbb{R}^{n}$ obtained by evaluating the activation-manifold spline at $B$ evenly spaced positions, $\mu_b = \gamma_{\mathcal{M}}(p_b)$. The mapping to behavior is
\begin{equation}
\label{eq:mc_belief_operator}
F(z) \;=\; \mathrm{softmax}\!\left( -\frac{\|z - \mu_b\|_2}{\tau} \right)_{b=1}^{B} \;\in\; \Delta^{B-1},
\end{equation}
with temperature $\tau = 0.5$. $F$ is a smooth, deterministic map from activations to position distributions. We use $B=128$, which makes $F$'s Jacobian full column-rank, ensuring the inverse problem has a locally unique solution. For each bin $i$, the natural centroid on the behavior manifold is the model's average output distribution conditioned on samples in that bin:
\begin{equation}
b_i \;=\; \mathbb{E}\!\left[ F(z) \mid \mathrm{bin}(z) = i \right] \;\in\; \Delta^{B-1}.
\end{equation}
Because the bin grid is dense relative to the data manifold's intrinsic dimension, $b_i$ is well-defined for all bins. We embed each $b_i$ in Hellinger coordinates $h_i = \sqrt{b_i}$ on the unit sphere of $\mathbb{R}^{B}$ and fit a 1D smoothing spline $\gamma_{\mathcal{M}_y} \colon \mathcal{Z} \to \mathbb{R}^{B}$ through $\{h_i\}$ parameterized by position. This is the behavior manifold $\mathcal{M}_y$.

\clearpage
\section{Additional Results}\label{app:results}

\subsection{In-Context Learning of Representations.}

In addition to results provided in the main text, we test a $9 \times 9$ cylinder in the ICLR domains, and find that despite the added complexity of a periodic dimension and substantially more graph nodes, when Llama 3.1 8B is provided sufficient context (2048 tokens in this case), it reaches above $80\%$ neighborhood accuracy (probability mass on valid neighbors). We fit a manifold and steer along this domain, finding that the result of factored control generalizes beyond the graph domain. 

\begin{figure}[h]
    \centering
    \includegraphics[width=1\linewidth]{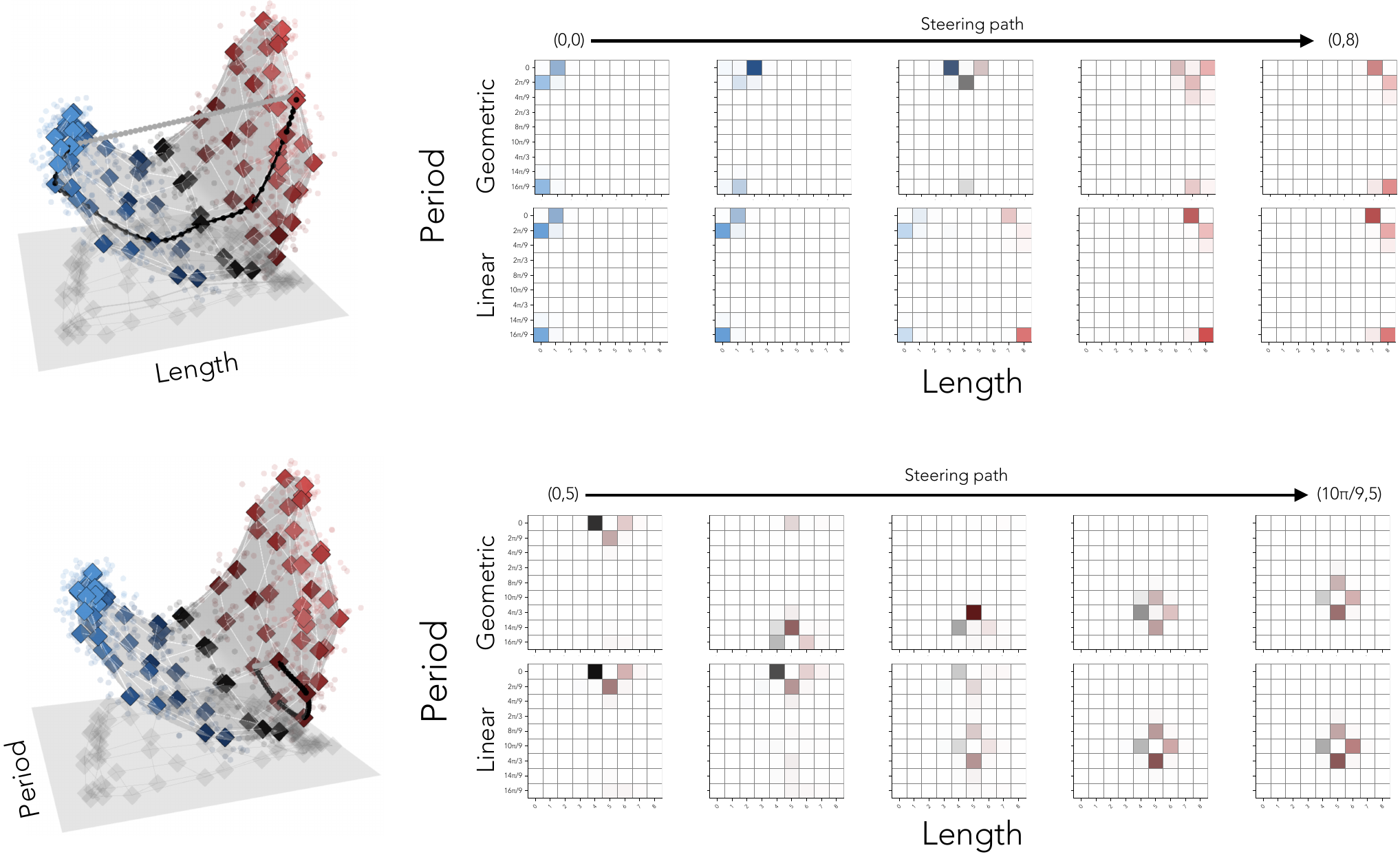}
    \caption{Results for in-context learning of representations on a $9\times9$ cylinder domain. We find that, as in the grid domain, manifold steering achieves factored control: coherent steering of independent dimensions, while linear steering once again shows `teleportation' behavior.}
    \label{fig:iclr-iclr-cylinder-real}
\end{figure}

\newpage 
\begin{figure}[h]
    \centering
    \includegraphics[width=1\linewidth]{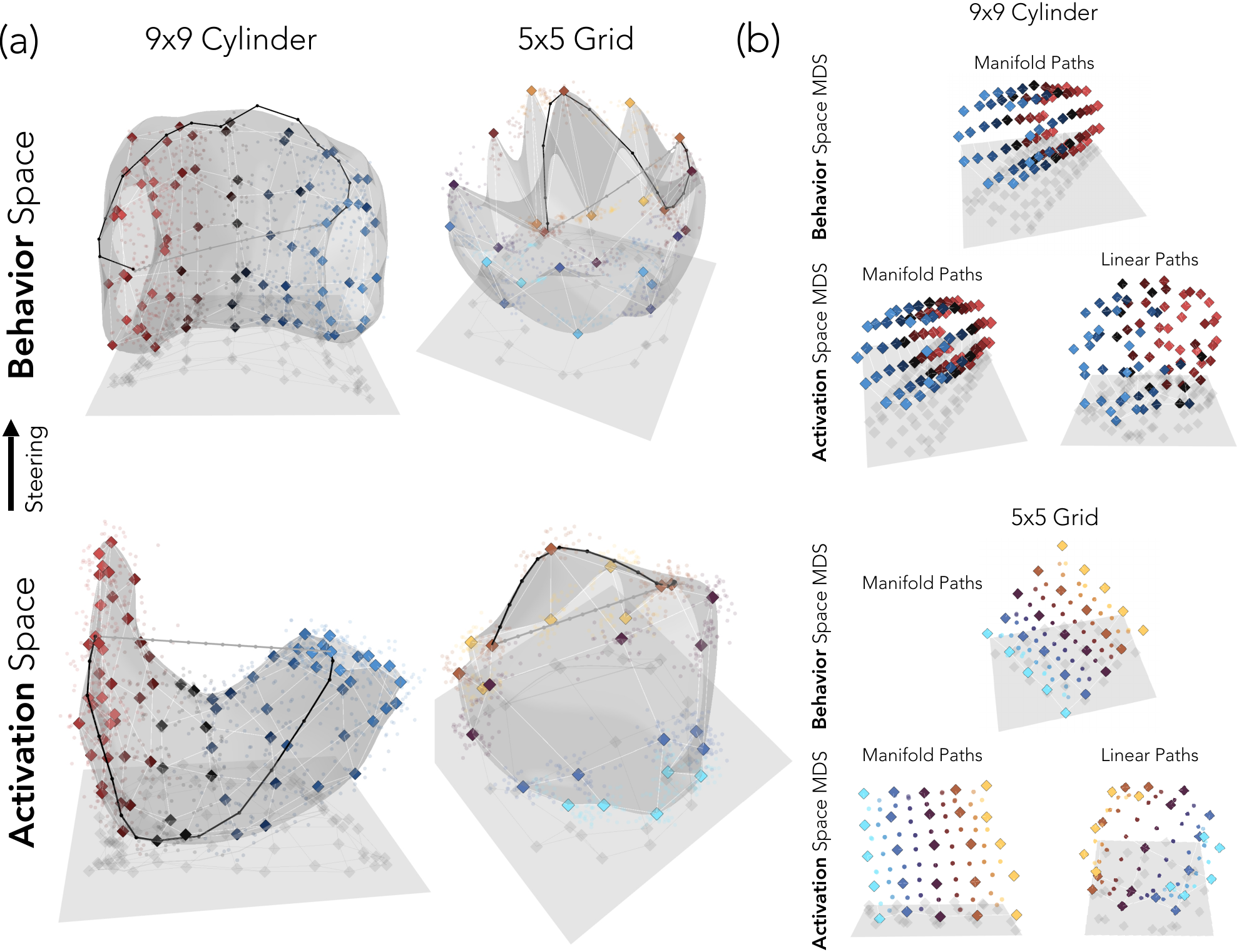}
    \caption{\textbf{(a)} Activation and behavior space paths for the $5\times5$ Grid task and $9\times9$ Cylinder. Similarly to the addition tasks with known concepts, we find that the manifold steering paths closely follow the behavior manifold $\mathcal{M}_y$. \textbf{(b)} Multidimensional scaling (MDS) embedding for linear and manifold distances in activation space and manifold distances in behavior space. As with the addition tasks with known concepts, manifold distances in activation space show a clear structural match to behavior space, whereas linear distances warp the structure. }
    \label{fig:iclr-iclr-cylinder-real}
\end{figure}

\newpage

\subsection{Manifold steering allows manipulation of uncertainty without loss of structure.}

\begin{figure}[h]
    \centering
    \includegraphics[width=\linewidth]{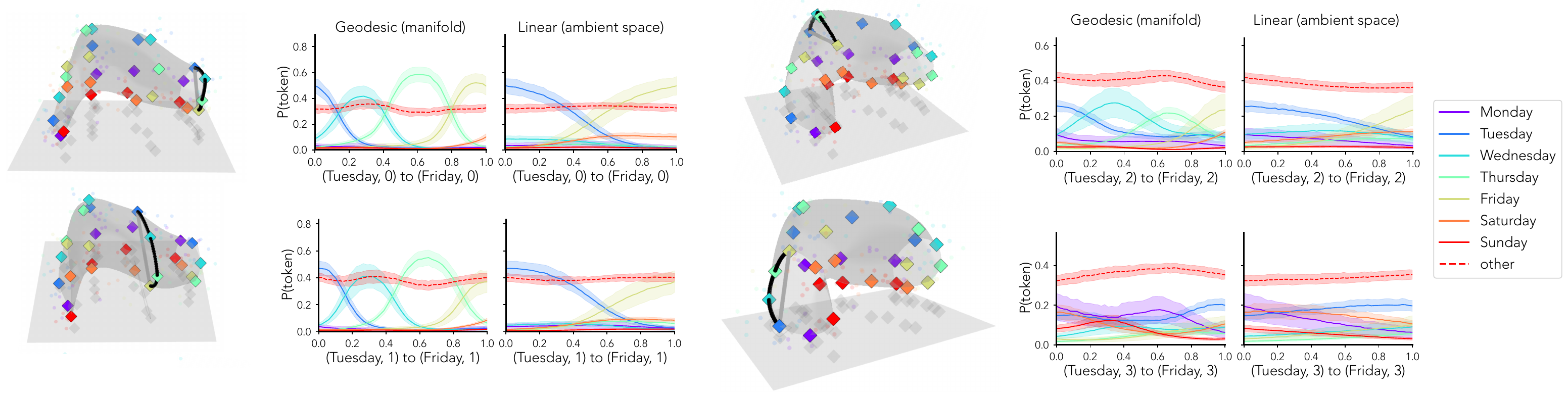}
    \caption{\textbf{Inducing greater uncertainty over a conceptual space and steering along it.} By increasing the addition value, we induce greater uncertainty in the model with respect to the right answer. Instead of grouping across addition values (as we do in Fig. \ref{fig:manifold-steering-main}), we visualize centroids by addition value, and find that these groups yield a series of circles organized into a curved cylinder-like shape. Manifold steering along the circle in the first three groups maintains ordered transitions, yet with increasing entropy in each group.}
    \label{fig:weekdays_uncertainty}
\end{figure}

To examine multi-dimensional concepts in known domains, we partition weekday addition centroids by addition value ($1$--$5$, $6$--$10$, $11$--$15$, $16$--$20$), revealing concentric circles along a second manifold dimension forming a cylinder-like structure (Fig.~\ref{fig:weekdays_uncertainty}). Manifold steering along the circular dimension maintains ordered weekday transitions with increasing entropy per group. This suggests manifold geometry can serve as a handle for calibrating model confidence in a controlled fashion. The experiment was conducted with Llama $3.1$ $70$B layer $70$. 

\newpage

\subsection{Mountain Car}
\label{sec:app_mountain_car}

\paragraph{Structural Correspondence between $\mathcal{M}_h$ and $\mathcal{M}_y$.}
If $\mathcal{M}_h$ encodes the model's predictive distributions over $\mathcal{Z}$, then $\mathcal{M}_h$ and $\mathcal{M}_y$ should be approximately isometric---distances along one manifold should correlate with distances along the other. We test this by sampling $W=50$ anchor positions inside the shared parameter range and computing pairwise arc lengths along each manifold:
\begin{equation}
d_{\mathcal{M}}(p_i, p_j) = \int_{p_i}^{p_j} \|\gamma_{\mathcal{M}_h}'(p)\|_2 \, dp,
\quad
d_{\mathcal{M}_y}(p_i, p_j) = \frac{1}{\sqrt{2}} \int_{p_i}^{p_j} \|\gamma_{\mathcal{M}_y}'(p)\|_2 \, dp,
\end{equation}
where the $1/\sqrt{2}$ on the behavior side converts the Euclidean integral in Hellinger ambient space to Hellinger units. The Pearson correlation between $\{d_{\mathcal{M}_h}(p_i, p_j)\}$ and $\{d_{\mathcal{M}_y}(p_i, p_j)\}$ over all $\binom{50}{2}=1225$ pairs is $r = \mathbf{0.996}$; the chord distances used by linear steering correlate far less ($r = 0.06$ between activation chord and behavior arc length), since chords cut across the encoder loop and are structurally divorced from the encoded conceptual geometry. 

\paragraph{Pullback: Behavior space steering.}

\begin{figure}[h]
    \centering
    \includegraphics[width=0.99\linewidth]{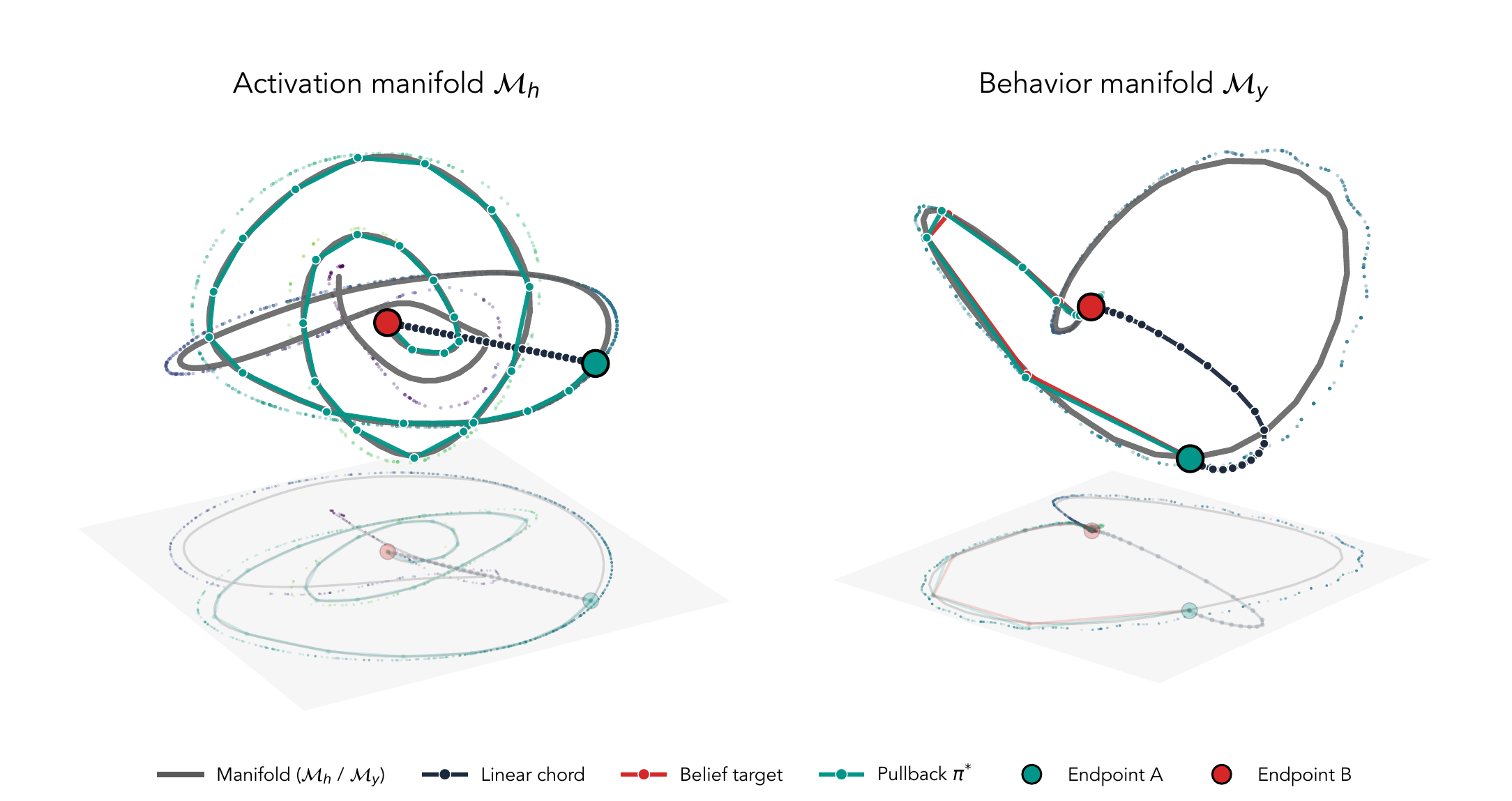}
    \caption{\textbf{Pullback from $\mathcal{M}_y$ recovers $\mathcal{M}_h$ in the visual world model.} \textbf{Left (Activation Space):} PCA visualization of the encoder representations, showing the geometric path along $\mathcal{M}_h$, the linear chord, and the pullback-optimized path $\pi^\star$ between endpoints $p_A$ and $p_B$. Although initialized at the chord, $\pi^\star$ converges onto $\mathcal{M}_h$, closing the spiral loop traced by the encoder geometry and becoming nearly indistinguishable from the activation reference. \textbf{Right (Behavior Space):} PCA visualization of the corresponding trajectories pushed through the operator $F$ (Eq.~\ref{eq:mc_belief_operator}), shown in Hellinger coordinates. The conformal target $\hat{\gamma}_\alpha$ tracks the behavior manifold $\mathcal{M}_y$, and the pushforward $F(\pi^\star)$ closely matches it, while the pushforward of the linear chord $F(\ell)$ departs sharply, cutting across the simplex interior rather than following $\mathcal{M}_y$. Together, the two panels show that optimizing an activation path to match a behavior-manifold target recovers $\mathcal{M}_h$ top-down: matching behavior along $\mathcal{M}_y$ is sufficient to pull activations back onto $\mathcal{M}_h$, mirroring the pullback result from the language-model experiments.}
    \label{fig:mc_pullback}
\end{figure} 
Having established the bottom-up direction in Fig.~\ref{fig:world_model_steering}, i.e., paths along $\mathcal{M}_h$ produce behavior trajectories on $\mathcal{M}_y$, We now test the top-down direction: starting from a behaviorally-natural trajectory in $\mathcal{M}_y$, do we naturally recover an activation path that traces $\mathcal{M}_h$? For each endpoint pair $(p_a, p_b)$ we construct the conformal behavior target $\hat{\gamma}_\alpha$ via the procedure of \S\ref{sec:act2} (geodesic on the simplex under cost $c(p) = \exp(\alpha \cdot d_H(p, \mathcal{M}_y))$). We then optimize an activation path $\pi_\alpha = (v_0, \dots, v_K)$ in $\mathbb{R}^{n}$ to minimize
\begin{equation}
\label{eq:mc_pullback_loss}
L(\pi) \;=\; \sum_{t=0}^{K} \bigl\| \sqrt{F(v_t)} - \sqrt{\hat{\gamma}_\alpha(t)} \bigr\|_2^2.
\end{equation}
Following the language-model setup, all $K+1$ waypoints (including endpoints) are free parameters, initialized at the linear chord and optimized jointly via L-BFGS with strong-Wolfe line search. We use $K=30$ waypoints and run independent optimizations for each of 30 endpoint pairs.

Across all 30 endpoint pairs, the pullback paths $\pi_\alpha$ closely trace $\mathcal{M}_h$ (Fig.~\ref{fig:mc_pullback}): The mean Euclidean distance from $\pi$ to $\mathcal{M}_h$, averaged over waypoints and pairs:
\[
\mathrm{linear\ chord}\!:\;\; 2.22,
\qquad
\mathrm{geometric}\;(\mathcal{M}_h)\!:\;\; 0.20,
\qquad
\mathrm{pullback}\!:\;\; 0.29.
\]
The pullback path is at $\mathbf{95.4\%}$ of the chord-to-geometric recovery and dominates the chord baseline on $30/30$ pairs. The aggregate degradation is concentrated on pairs with one endpoint at the extreme wall position ($p \approx -1.2$), where the encoder geometry has tighter curvature; on the remaining $\sim$20 pairs, $\pi_\infty$ is essentially indistinguishable from $\mathcal{M}_h$ itself. The $\alpha$-sweep traces the same family of trajectories observed in the language-model experiments: at $\alpha = 0$ the conformal target is the unrestricted Hellinger geodesic on the simplex, and the recovered $\pi_0$ leaves $\mathcal{M}_h$ in order to match this off-manifold target; as $\alpha$ grows the target is pushed onto $\mathcal{M}_y$ and the recovered $\pi_\alpha$ correspondingly tracks $\mathcal{M}_h$.

\end{document}